\def\PAGENUMS{1}
\def\AAM{1}

\documentclass[letterpaper]{article} % DO NOT CHANGE THIS
\usepackage{aaai2026}  % DO NOT CHANGE THIS
\usepackage{times}  % DO NOT CHANGE THIS
\usepackage{helvet}  % DO NOT CHANGE THIS
\usepackage{courier}  % DO NOT CHANGE THIS
\usepackage[hyphens]{url}  % DO NOT CHANGE THIS
\usepackage{graphicx} % DO NOT CHANGE THIS
\usepackage{natbib}  % DO NOT CHANGE THIS AND DO NOT ADD ANY OPTIONS TO IT
\usepackage{caption} % DO NOT CHANGE THIS AND DO NOT ADD ANY OPTIONS TO IT
\usepackage{algorithm}
\usepackage{algorithmic}
\usepackage{amsmath}
\usepackage{amssymb}
\newtheorem{theorem}{Theorem}
\newtheorem{corollary}[theorem]{Corollary}

\newtheorem{lemma}[theorem]{Lemma}
\usepackage{graphicx,tabularx,booktabs}
\usepackage{subcaption}
\usepackage{caption}

\usepackage{siunitx} % for S column type
\usepackage{etoolbox} % for underlining in S columns
\usepackage{fancyhdr} % for AAM headers
\usepackage{lastpage} % for pre-print page numbers

\usepackage{newfloat}
\usepackage{listings}
\DeclareCaptionStyle{ruled}{labelfont=normalfont,labelsep=colon,strut=off} % DO NOT CHANGE THIS
\floatstyle{ruled}
\newfloat{listing}{tb}{lst}{}
\floatname{listing}{Listing}
\nocopyright

\title{How Wide and How Deep? Mitigating Over-Squashing of GNNs \\ via Channel Capacity Constrained Estimation}
\author{
    Zinuo You\textsuperscript{\rm 1}, Jin Zheng\textsuperscript{\rm 2}, John Cartlidge\textsuperscript{\rm 2}
}
\affiliations{
    \textsuperscript{\rm 1}School of Computer Science\\
    \textsuperscript{\rm 2}School of Engineering Mathematics and Technology\\
    University of Bristol, UK\\
    zinuo.you@bristol.ac.uk, jin.zheng@bristol.ac.uk, john.cartlidge@bristol.ac.uk
}

\begin{document}

\maketitle
% --- Force numbering of sections/subsections ---
\setcounter{secnumdepth}{2}  % 1 = sections only, 2 = sections + subsections, etc.
\renewcommand{\thesection}{\arabic{section}}
\renewcommand{\thesubsection}{\thesection.\arabic{subsection}}

\begin{abstract}
Existing graph neural networks typically rely on heuristic choices for hidden dimensions and propagation depths, which often lead to severe information loss during propagation, known as over-squashing. To address this issue, we propose Channel Capacity Constrained Estimation (C$^3$E), a novel framework that formulates the selection of hidden dimensions and depth as a nonlinear programming problem grounded in information theory. Through modeling spectral graph neural networks as communication channels, our approach directly connects channel capacity to hidden dimensions, propagation depth, propagation mechanism, and graph structure. Extensive experiments on nine public datasets demonstrate that hidden dimensions and depths estimated by C$^3$E can mitigate over-squashing and consistently improve representation learning. Experimental results show that over-squashing occurs due to the cumulative compression of information in representation matrices. Furthermore, our findings show that increasing hidden dimensions indeed mitigates information compression, while the role of propagation depth is more nuanced, uncovering a fundamental balance between information compression and representation complexity.
\end{abstract}

\begin{links}
    \link{Code}{https://github.com/pixelhero98/C3E}
    % \link{Datasets}{https://aaai.org/example/datasets}
    % \link{Extended version}{https://aaai.org/example/extended-version}
\end{links}

%% Add page numbers - for pre-prints only
\ifnum\PAGENUMS=1
    \thispagestyle{fancy}
    \pagestyle{fancy}
    \fancyfoot[C]{\fontsize{9}{10} \selectfont Page \thepage ~of \pageref{LastPage}}% add page number (smaller font size defined)
    \fancyhead[L,C,R]{} % clear header text
    
    % Move header line and text higher
    \setlength{\topmargin}{-10mm}
    \setlength{\headheight}{15pt}
    \setlength{\headsep}{8pt}

    \ifnum\AAM=1 
        \fancyhead[C]{\fontsize{9}{10} 
        \selectfont Accepted author manuscript: You, Zheng, \& Cartlidge (2026), 40th Annual AAAI Conference on Artificial Intelligence (AAAI-26)}
    \else
        \renewcommand{\headrulewidth}{0pt} % no header text, so removing horizontal line in header
    \fi
\fi

%--- MAIN BODY ---%
% Main body of text for AAAI-26: How Wide and How Deep? Mitigating Over-squashing of GNNs via Channel Capacity Constrained Estimation. You, Zheng, and Cartlidge

\section{Introduction}
\label{intro}
Graph Neural Networks (GNNs) have emerged as powerful tools in various graph-related learning tasks~\citep{bruna2014spectral,defferrard2016convolutional}, which stem from the goal of learning meaningful representations over graphs. Broadly, GNNs can be categorized into spectral and spatial methods. The spatial GNNs propagate information in the spatial domain, such as node-wise or edge-wise operations~\citep{hamilton2017inductive,xu2018powerful,lee2019self} and feature-dependent operations~\citep{xu2018representation,velivckovic2018graph,rampavsek2022recipe}. Conversely, spectral GNNs~\citep{defferrard2016convolutional,kipf2017semi,he2021bernnet,wang2022powerful} are grounded in spectral graph theory, where propagation methods are grounded on spectral filters. Moreover, spectral GNNs offer transparent, traceable propagation over graphs that can be expressed in closed form via matrix operations, unlike most spatial GNNs. Despite these merits, GNNs suffer a fundamental drawback from increasing propagation depth, which leads to severe performance degradation. This phenomenon is commonly attributed to over-smoothing \citep{nt2019revisiting,cai2020note} or over-squashing~\citep{alon2020bottleneck,topping2021understanding,di2023over,huang2024universal}. The former refers to node representation becoming overly similar during propagation, and the latter refers to information in node representation being severely squashed into limited-size vectors during propagation, leading to catastrophic information loss in the learned representation~\citep{alon2020bottleneck,topping2021understanding,di2023over}. 

While over-smoothing has been extensively studied and can be mitigated with various normalization and residual techniques~\citep{nt2019revisiting,cai2020note,chen2020measuring,bodnar2022neural,maskey2024fractional}, over-squashing remains less understood. Existing works have largely focused on modifying graph structures or attribute features to mitigate over-squashing. For instance, graph rewiring techniques (e.g., adding or dropping edges~\citep{topping2021understanding}) and spectral methods (e.g., spectral graph rewiring~\citep{karhadkar2022fosr} or modifying edges via spectral gap optimization~\cite{gravina2025oversquashing}). However, a growing body of theoretical works,~\citet{loukas2020hard,di2023over}, demonstrate that realizing effective information propagation in GNNs is not solely a function of propagation methods and graph structures; it is critically dependent on network architectures, i.e., propagation depths, and hidden dimensions. For example,~\citet{loukas2020graph} points out that existing GNNs fail to effectively propagate information unless the product of their propagation depths and hidden dimensions exceeds a certain polynomial related to the graph size. Similarly, \citet{di2023over} further show that larger hidden dimensions can mitigate over-squashing, and specific propagation depths can be helpful for representation learning before leading to vanishing gradients. These align with prior empirical analysis \citep{yang2020revisiting,cong2021provable,zhou2021understanding} that such degradation arises mainly from over-simplifying learnable matrices in analysis. While prior works show that specific hidden dimensions and depth can mitigate over-squashing, they offer no method to obtain these parameters. In modern representation learning, graph features propagated via message passing are subsequently transformed by learnable weight matrices, producing fixed-size representation embeddings~\citep{dwivedi2023benchmarking}. From an information-theoretic perspective, over-squashing essentially corresponds to information loss in these learned representation matrices. Since entropy measures a variable’s uncertainty or information content~\citep{jaynes1957information,kullback1997information}, it serves as a natural choice for quantifying the information retained in representation matrices. This provides an explicit way to track the information flow propagated through the network~\citep{saxe2019information,wu2020graph,shen2023deepmad}, rather than inferring it indirectly from graph structures alone, like former remedies. 

In this paper, we propose the \textbf{Channel Capacity Constrained Estimation} (C$^3$E), a novel theoretical framework for estimating optimal hidden dimensions and propagation depth for spectral GNNs before training. Although knowing the exact state of the network before training is impossible, we can leverage the principle of maximum entropy~\citep{jaynes1957information} to estimate an upper bound on the information that a spectral GNN can propagate. Moreover, by invoking Shannon’s Theorem~\citep{shannon1948mathematical}, we note that a communication channel can achieve near error-free information transmission with a certain encoding scheme, if its capacity exceeds the information load. As former studies~\citep{bruna2014spectral,henaff2015deep,defferrard2016convolutional} suggest, learnable matrices can be interpreted as encoders of graph signals, affecting the encoded representation of the network. Similarly, we model a spectral GNN as a communication channel whose capacity depends on hidden dimensions and propagation depth. Under this perspective, estimating optimal width and depth reduces to maximizing channel capacity under Shannon’s Theorem, yielding a nonlinear programming formulation. The contributions of this work are threefold. First, we provide an information-theoretic view to model information flow in spectral GNNs, which links hidden dimensions, depth, propagation methods, and graph structures to the encoded representation. Second, we formulate a nonlinear programming problem to estimate optimal hidden dimensions and propagation depth, providing a principled approach to choosing network architectures. Third, we demonstrate that optimal hidden dimensions and propagation depth derived from C$^3$E effectively mitigate over-squashing and consistently improve representation learning, without altering propagation methods or graphs.

\section{Related work}
\label{sec2}
\subsection{Information Theory in Representation Learning}
Information theory~\citep{shannon1948mathematical,jaynes1957information,kullback1997information} has long been a powerful tool for analyzing neural networks. For instance,~\citet{saxe2019information} have explored the entropy distribution and the information flow of learned representation in deep neural networks. It reveals that the information compression in representation matrices occurs along with the increase in network depth. The information bottleneck principle is broadly applied in neural networks to learn minimal effective representations, which maximize the mutual information between the learned representation and the target to alleviate potential information loss~\citep{tishby2000information,wu2020graph}. Furthermore,~\citet{sun2021mae} have managed to generate highly competitive deep convolutional neural networks (CNNs) based on the principle of maximum entropy~\citep{jaynes1957information,jaynes2003probability}. Recently, some studies~\citep{chan2022redunet,roberts2022principles} endeavor to establish relationships between entropy and representation matrices of neural networks. 

\subsection{Spectral GNNs}
Spectral-based GNNs define propagation using spectral filters or kernel functions. Much of the progress in the field, from early models like GCN~\citep{kipf2017semi}, APPNP~\citep{gasteiger2018predict}, and SGC~\citep{wu2019simplifying}, to advanced methods like GDC (two variants: GDC\textsubscript{HK} and GDC\textsubscript{PPR})~\citep{gasteiger2019diffusion}, GPRGNN~\citep{chien2020adaptive}, S$^2$GC~\citep{zhu2021simple}, ChebyNetII~\citep{he2022convolutional}, and JacobiConv~\citep{wang2022powerful}. has focused on designing sophisticated filters to overcome over-smoothing. While successful, this focus on the propagation mechanism has revealed a more fundamental and architectural bottleneck: over-squashing. This form of information loss stems not from the filter but from the capacity of learnable weight matrices that process propagated signals. As recent studies~\citep{zhou2021understanding,cong2021provable,di2023over} confirm, the dimensions and depth of learnable transformations are critical factors, yet their choice has often been overlooked.

\section{Preliminary}
\label{sec3}
\subsection{Entropy of Matrix}
We define the entropy of a real-valued matrix $\mathbf{Z}$ by treating its entries as samples from a random variable $ Z\sim p$. This entry‐wise definition serves as a tractable proxy for the matrix's total information content by abstracting away higher-order correlations between entries (detailed \textbf{justification} and \textbf{discussion} are provided in Appendix~\ref{matenapp}). If the latent distribution $p$ is continuous, then its entropy is given by,
\begin{equation}
H(Z) = -\int_{-\infty}^{\infty} p(z)\ln{(p(z))}d z.
\label{repen}
\end{equation}
For the given mean $\mu_{Z}$ and variance $\sigma^{2}_{Z}$, it is maximized by a Gaussian distribution $\mathcal{N}(\mu_{Z},\sigma^{2}_{Z})$, which yields, 
\begin{equation}
    H(Z) \leq\frac{1}{2} \ln{(2 \pi \mathrm{e} \sigma^2_{Z})}.
    \label{realmax}
\end{equation}
Yet, we do not have access to the true underlying distribution $p$ a priori. Instead, we have the finite-dimensional matrix $\mathbf{Z}\in \mathbb{R}^{\alpha \times \beta}$, which constitutes a finite set $\alpha \beta$ samples. These samples form an empirical distribution whose information content is captured by the discrete form,
\begin{equation}
H(Z) = -\sum_{i=1}^{|\mathrm{Supp}(Z)|} P(Z = z_i) \ln{(P(Z = z_i))},
\label{discreteen}
\end{equation}
where $\text{Supp}(\cdot)$ denotes the support set, i.e., the set of all possible values $\{z_i\}$ that entries of $\mathbf{Z}$ can take. Then, the maximum entropy is bounded by,
\begin{equation}
    H(Z) \leq \ln{(|\mathrm{Supp}(Z)|)}\leq \ln{(\alpha\beta)}.
    \label{discreteenmax}
\end{equation}
The \textbf{proofs} of Eq.~(\ref{realmax}) and Eq.~(\ref{discreteenmax}) are provided in Appendix~\ref{proofenup}. These show that the maximum information a matrix can convey is capped by its dimensionality. A lower-dimensional matrix inherently has a smaller maximum support size, imposing a coarser discretization on the representation of any underlying distribution.

\subsection{Entropy of Graph}
Graph entropy quantifies the uncertainty or information content of a graph $\mathcal{G}$. While definitions vary, they generally measure the information content of the graph based on some property extraction function $g(\cdot)$ (e.g., eigenvector centrality or homophily/heterophily metrics). A generalized form of graph entropy~\citep{dehmer2011history} is,
\begin{equation}
    H(\mathcal{G}) = -\sum_{i=1}^{n}\frac{g(v_i)}{\sum_{j=1}^{n}g(v_j)}\ln{\frac{g(v_i)}{\sum_{j=1}^{n}g(v_j)}}.
    \label{gendef}
\end{equation}
Here, $v_i$ denotes the $i$-th vertex, and $n$ denotes the number of nodes.

\subsection{Channel Capacity}
In terms of information theory~\citep{shannon1948mathematical,jaynes1957information,gallager1968information}, channel capacity is defined as the theoretical maximum of which information can be reliably transmitted over a communication channel. The channel capacity of a communication channel is expressed as~\citep{shannon1948mathematical},
\begin{equation}
    \phi =  \max\;I(f(\mathcal{M});\mathcal{M}').
    \label{cpdef}
\end{equation}
Here, $I(f(\mathcal{M});\mathcal{M}') = H(f(\mathcal{M})) - H(f(\mathcal{M})|\mathcal{M}')$ denotes the mutual information between encoded input $f(\mathcal{M})$ and the output $\mathcal{M}'$, and $f(\cdot)$ denotes the encoder.

\section{Methodology}
\subsection{Theoretical Channel Capacity of Spectral GNNs}
Drawing on prior research~\citep{saxe2019information,sun2021mae,chan2022redunet,shen2023deepmad,yang2023minimum} about information flows propagated in neural networks, we extend the classical definition of channel capacity to GNNs, which we model as communication channels. To establish a formal theoretical framework grounded in this perspective, a class of models with analytical tractability is required. Spectral GNNs provide the ideal characteristic, as their propagation mechanisms can be collapsed into a single matrix operator.

Consider a spectral GNN with $L$ propagation layers learning representations on a graph $\mathcal{G}$. We collapse the propagation operation on the adjacency matrix $\mathbf{A}\in \mathbb{R}^{n \times n}$ into a propagation matrix $\mathbf{S}_l \in \mathbb{R}^{n \times n}$. Then, its layer-wise representation learning operation becomes,
\begin{equation}
    \mathbf{H}_{l} = \Delta(\mathbf{S}_l\mathbf{H}_{l-1}\mathbf{W}_{l}).
    \label{gnndefinition}
\end{equation}
Here, $\mathbf{H}_l \in \mathbb{R}^{n\times w_{l}}$ denotes the latent representation matrix, $\mathbf{H}_0 \in \mathbb{R}^{n \times m}$ denotes the initial feature matrix ($m=w_0$), $\mathbf{W}_l \in \mathbb{R}^{w_{l-1} \times w_{l}}$ denotes the learnable weight matrix, and $\Delta(\cdot)$ denotes the nonlinear activation function. Therefore, the encoded representation for the downstream task is $\mathbf{H}_L \in \mathbb{R}^{n \times w_{L}}$. In addition, the above layer-wise propagation framework directly extends to models like SGC and APPNP, whose underlying spectral formalisms provide a collapsible operator that is computed via a spatial message-passing scheme.

If the GNN encoder $f(\cdot)$ learns the exact mapping $f:\mathcal{G}\to\mathcal{G}'$, where $\mathcal{G}'$ is the graph $\mathcal{G}$ with node representations that are sufficient to resolve uncertainty about downstream tasks (e.g., semi-supervised node classification or property prediction), then the conditional entropy ${H}(\mathcal{G}'|f(\mathcal{G}))$ vanishes:
\begin{equation}
    {H}(\mathcal{G}'|f(\mathcal{G}))=0.
    \label{conditionen}
\end{equation}
This means the encoded representation $f(\mathcal{G}) = \mathbf{H}_L$ perfectly matches $\mathcal{G}'$ with zero uncertainty. Given our lack of prior knowledge about network states in advance, then, by the principle of maximum entropy, we should select the probability distribution that best represents the system's current state, which is the one with the largest entropy, subject to known constraints~\citep{jaynes1957information,jaynes2003probability}. Consequently, combining this crucial premise with Eq.~(\ref{realmax}), Eq.~(\ref{cpdef}), Eq.~(\ref{gnndefinition}), and Eq.~(\ref{conditionen}), we arrive at the following Theorem.
\begin{theorem}
The channel capacity of a spectral GNN is defined by maximizing the entropy of the encoded representation $\mathbf{H}_L$, which is expressed as,
\begin{align}
    \phi & = \underset{}{\max}\;I(f(\mathcal{G});\mathcal{G}')\notag\\
    & = \underset{}{\max}\; H(f(\mathcal{G})) - {H}(f(\mathcal{G})|\mathcal{G}')\notag\\
    & = \max\;H(\mathbf{H}_{L}) \notag\\
    & = \max\;\left[\frac{1}{2}\ln{(2\pi \mathrm{e})} + \frac{1}{2}\sum_{l=1}^{L}\ln{(nw_{l-1}\sigma^2_{\mathbf S_{l}})}\right]. 
    \label{gnnccdef}
\end{align}
\label{theo1}
\end{theorem}
The \textbf{proof} is provided in \ref{proof4.1}. Here, a larger channel capacity corresponds to greater information content that can be carried by the network, indicating that the network can represent a more complex and informative distribution. The formula shows that hidden dimensions and propagation depth are crucial to the channel capacity. The variance term $n\sigma^2_{\mathbf S_{l}}$ depicts the role of graph structures and propagation mechanisms, where larger hidden dimensions amplify their effects and smaller ones diminish them. Moreover, if hidden dimensions are too small, it results in reductions in channel capacity and information loss in the encoded representation as the propagation depth increases. This aligns with theoretical analysis from previous studies \citep{loukas2020hard,di2023over} that the product of hidden dimensions and propagation operator should be sufficiently large to avoid information loss.

\subsection{Information Compression and Over-Squashing}
Recently, some studies \citep{topping2021understanding,di2023over,huang2024universal} have attempted to measure the information compression in GNNs via the graph bottleneck or the Jacobian of latent representation. For instance,~\citet{topping2021understanding} illustrate that the graph bottleneck leads to severe information compression and hence over-squashing, originating from high negative curvature edges. According to~\citet{saxe2019information}, information compression occurs along with the representation learning process. Furthermore, other studies~\citep{di2023over,huang2024universal} show that over-squashing is closely related to choices of hidden dimensions and propagation depth. However, these measures are hard to obtain due to complex graph structures and variations in training. Thus, we introduce a metric termed representation compression ratio to measure the severity of information compression,
\begin{equation}
    \theta = \frac
    {\phi}{\overline{w}},\;\overline{w}=(\prod_{l=1}^{L} w_l)^{\frac{1}{L}}.
    \label{theta}
\end{equation}
Here, $\overline{w}$ denotes the geometric mean of hidden dimensions, serving as an equivalent representation dimension per layer. The metric $\theta$ provides an explicit measure of the information compression intensity in the representation matrix. If $\theta \rightarrow \infty$, i.e., $\phi \gg \overline{w}$, then this implies that per equivalent representation dimension compresses massive information content severely. If $\theta \rightarrow 0$, i.e., $\overline{w} \gg \phi$, then this means that the representation dimension is over-provisioned relative to the information content. 

To better understand the specific effects of hidden dimensions and propagation depth on information compression, we analyze based on the representation compression ratio $\theta$. Substituting Eq.~(\ref{gnnccdef}) into Eq.~(\ref{theta}) and deriving the partial derivatives yields the following Corollary.
\begin{corollary}
The representation compression ratio $\theta$ of a spectral GNN exhibits a dual dependency on $\overline{w}$ and $L$. Let $\overline{K} = \mathbb{E}_l[\ln{(n\sigma^2_{\mathbf S_l})}]$ be the average log propagation variance of graph structures and propagation operations.
\begin{itemize}
    \item On $\overline{w}$: Given fixed $L$, $\theta$ is maximized at a threshold $\overline{w}^*$, where $\theta$ monotonically increases for $0 < \overline{w} \leq \overline{w}^*$ and monotonically decreases for $\overline{w} > \overline{w}^*$. The threshold $\overline{w}^* =\mathrm{e}^{\left[1 - \frac{\ln{(2\pi\mathrm{e})}+\ln{(m)}-\ln{(w_L)}+\sum_{l=1}^L\ln{(n\sigma^2_{\mathbf S_l})}}{L}\right]}$ eventually converges to $\underset{L \rightarrow\infty}{\lim}\overline{w}^* = \mathrm{e}^{1-\overline{K}}$.
    \item On $L$: Given fixed $\overline{w}$, the effect of increasing $L$ depends on $\overline{w}$ relative to properties of the graph and propagation method. If $\ln{(\overline{w})}>-\overline{K}$, increasing $L$ increases $\theta$, exacerbating information compression. Conversely, when $\ln{(\overline{w})}\leq -\overline{K}$ increasing $L$ directly decreases $\phi$ and declines $\theta$, symptomatic of information loss.
\end{itemize}
\label{theo2}
\end{corollary}
The \textbf{proof} is provided in \ref{col2}. First, expanding hidden dimensions is the primary choice for mitigating high information compression (over-squashing). Second, better propagation methods or graph rewiring indeed mitigate such information compression by decreasing $\overline{w}^*$ and increasing $\overline{K}$. Nevertheless, the role of propagation depth is conditional and reveals two failure modes: deep and wide GNNs might suffer from over-squashing due to cumulative information compression ($\theta$ increases when $\overline{w}\leq\overline{w}^*$), while deep and narrow GNNs suffer from over-squashing (since the propagated information $\phi$ vanishes when $\ln{(\overline{w})}<-\overline{K}$). These results in Corollary~\ref{theo2} align with former empirical findings~\citep{loukas2020graph,topping2021understanding,di2023over}. 

\subsection{From Theoretical Limit to Effective Channel Capacity}
Previously, Theorem~\ref{theo1} establishes the theoretical channel capacity of GNNs, which offers a critical upper bound on the information the network can encode. Nevertheless, this global perspective treats the network as a whole system, which does not explicitly consider the architectural constraints imposed by the layer-by-layer information propagation~\citep{achille2018emergence,saxe2019information}. In practice, a very narrow layer following a very wide one will structurally cap the information passed to subsequent layers.

To fill this gap, we introduce the effective channel capacity $\phi_0$, which accounts for architectural constraints between adjacent layers. By modeling each learnable weight matrix transformation as a communication channel, whose structure is analogous to a complete bipartite graph between its input and output neurons~\citep{pellizzoni2024expressivity}, the effective channel capacity $\phi_0$ is defined by the following expression,
\begin{equation}
\phi_0 = \sum_{l=1}^{L}  \frac{\ln{\left(\frac{w_{l-1} w_l}{w_{l-1} + w_l}\right)}}{\frac{\ln{(2\pi \mathrm{e})}}{\ln{(nw_{l-1}\sigma^2_{\mathbf S_l})}} + \sum_{o=1}^{l}\frac{\ln{\left(nw_{o-1} \sigma^2_{\mathbf S_{o}}\right)}}{\ln{(nw_{l-1}\sigma^2_{\mathbf S_l})}}}.
\label{cclower}
\end{equation}
The \textbf{justification} is provided in \ref{theo6.1}. This expression provides a practical and architecture-aware measure of channel capacity by capturing two fundamental dynamics. First, the numerator models architectural bottlenecks; it shows that large disparities between the widths of adjacent layers structurally reduce the information that can be retained. Second, the denominator reflects a cumulative attenuation effect, where the information from preceding layers (including initial features) diminishes the relative contribution of the current layer. Together, these terms formalize that to preserve sufficient information capacity, besides propagation mechanisms and graph structures, GNNs should avoid sharp changes in hidden dimensions and that deeper layers provide diminishing returns on the capacity. These findings provide a principled framework for empirical results observed in prior work~\citep{loukas2020hard,loukas2020graph,cong2021provable,di2023over}. 

\begin{figure*}[t]
    \centering
    \includegraphics[width=0.95\linewidth]{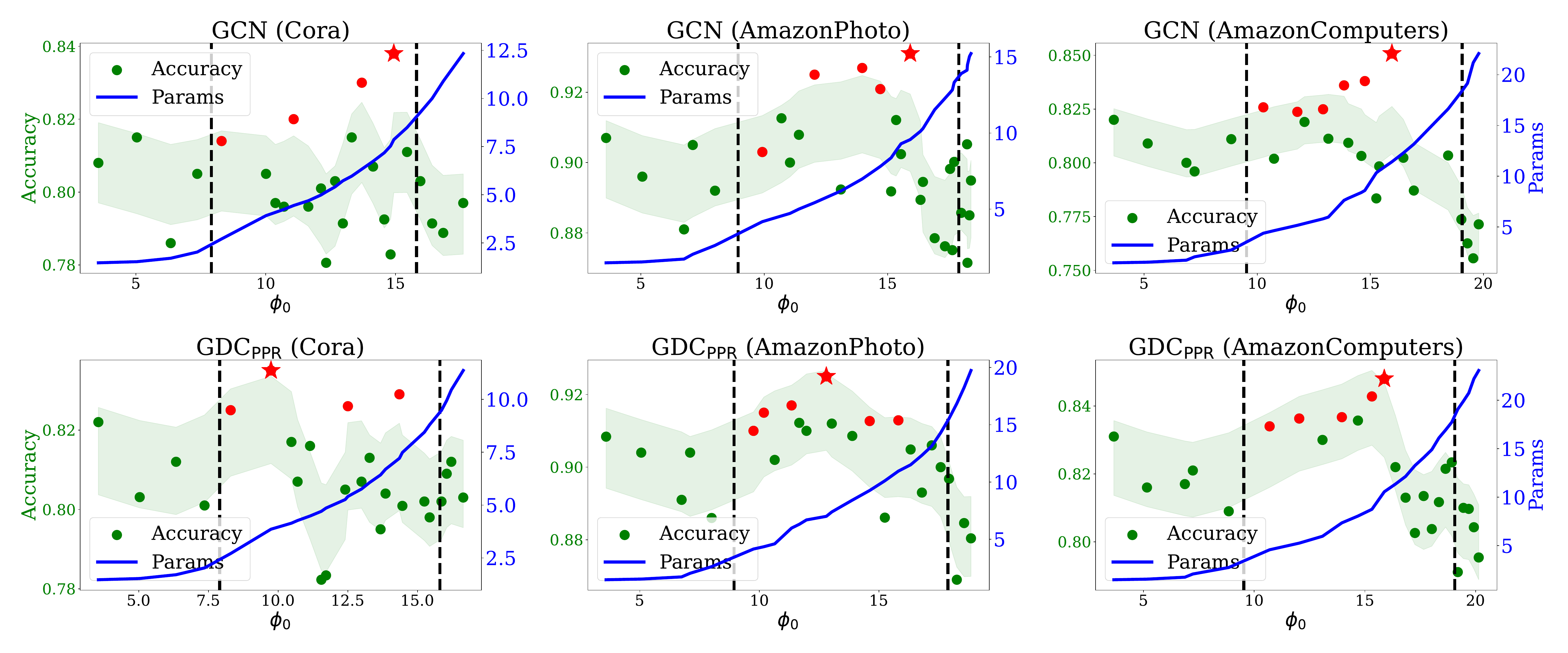}
    \caption{The green-axis (\textbf{left}) denotes performance, and blue-axis (\textbf{right}) denotes parameters counts (in millions). C$^3$E-estimated solutions (red points, \textbf{starred} for optimal) consistently land in high-performance regions within \textbf{dashed} intervals defined in Eq.~(\ref{lbcons}), outperforming baselines with heuristic dimensions (green points, e.g., 16, 32, 64, 128, 256, 512, 1024).}
    \label{effex_analysis}
\end{figure*}
\begin{table*}[ht]
  \centering
    \setlength{\tabcolsep}{1mm}
    \begin{tabular}{lccccccccc}
      \toprule
      Statistics   & Cora & Citeseer & Pubmed & AmazonPhoto & AmazonComputers & Chameleon & Squirrel & ogbn-arxiv & ogbn-papers100M \\
      \midrule
      \# Node      & 2,708 & 3,327     & 19,717  & 7,650        & 13,752           & 2,277      & 5,201  &  169,343 & 111,059,956 \\
      \# Feature   & 1,433 & 3,703     & 500    & 745         & 767             & 2,325      & 2,089   &  128 & 128\\
      \# Edges   & 5,429 & 4,732     & 44,338    & 119,043         & 245,778             & 36,101      & 217,073   &  1,166,243 & 1,615,685,872\\
      \# Classes   & 7 & 6     & 3    & 8         & 10             & 5      & 5   &  40 & 172\\
    Avg Time         & 3.7  & 4.9      & 5.9    & 5.3         & 5.7             & 3.6       & 5.0    & 239.2 & 879.6 \\
      \bottomrule
    \end{tabular}
      \caption{Statistics of the experimental datasets, including average time (in seconds) for C$^3$E to generate solutions for baselines.}
        \label{time}
\end{table*}

\subsection{Channel Capacity Constrained Estimation}
The preceding analysis demonstrates the need to balance channel capacity against the risk of information compression. Accordingly, our estimation should effectively manage this trade-off. First, Shannon's Theorem~\citep{shannon1948mathematical} states that for near error-free information propagation, the channel capacity must meet or exceed the information being transmitted: $\phi \ge H(\cdot)$. Since $\mathcal{G'}$ is unknown a priori, we utilize the maximum possible graph entropy, $H_{\max}(\mathcal{G}) = \ln(n)$, as a safe lower bound for the required channel capacity. This gives the first condition: $\phi_0 \ge \ln(n)$. Second, as established in Corollary~\ref{theo2}, simply maximizing channel capacity is not ideal, as it can lead to over-squashing by excessive information compression. To prevent this, we introduce a hyperparameter $\eta \in (0, 1]$ to regularize the effective channel capacity, imposing an upper bound on $\phi_0$. This gives the second condition: $\phi_0 \le \frac{1}{\eta}\ln(n)$. Then, the two conditions form the effective trade-off,
\begin{equation}
    \ln{(n)}\leq \phi_0 \leq \frac{1}{\eta} \ln{(n)}.
    \label{lbcons}
\end{equation}
This constraint ensures the GNN has sufficient channel capacity and prevents runaway information compression. We treat $\eta$ as a tunable regularizer. Our sensitivity analysis in \ref{etarel} demonstrates, the framework remains robust across a wide range of $\eta$ values; tuning up $\eta$ speeds solutions via fewer candidate solutions, and vice versa. %suggesting that delicate fine-tuning is not required.

Putting Theorem~\ref{theo1}, Corollary~\ref{theo2}, Eq.~(\ref{lbcons}) together, the C$^3$E is expressed by the following form,
%the C$^3$E framework seeks to maximize the theoretical channel capacity subject to the constraints obtained from our analysis of information compression and effective channel capacity,
%. 
\begin{align}
    &\max_{\mathbf{w}^{(L)}, L} \;\left[\frac{1}{2}\ln{(2\pi \mathrm{e})} + \frac{1}{2}\sum_{l=1}^{L}\ln{(nw_{l-1}\sigma^2_{\mathbf S_{l}})}\right]  \label{programming}\\
    &\;s.t. \;\;\; \quad  \overline{w} > \overline{w}^*,\;\ln{(\overline{w})} > -\overline{K},\notag\\
    & \;\quad\quad \quad   \ln{(n)}\leq \phi_0 \leq \frac{1}{\eta}\ln{(n)}. \notag
\end{align}
Here, $\mathbf{w}^{(L)} = \{w_1,..., w_L\}$. The primary objective seeks to maximize the theoretical channel capacity. The first constraint prevents two failure modes: either excessive information compression or information loss as identified in Corollary~\ref{theo2}. The second constraint enforces the GNN to possess sufficient channel capacity while avoiding the pitfalls of naively increasing $L$ or $\overline{w}$. These constraints steer trivial solutions of infinitely large width or depth away and instead force a trade-off between propagation depth and hidden dimensions. Feasible solutions can be obtained using off-the-shelf solvers for constrained nonlinear programming, such as SLSQP \citep{kraft1988software}. For implementation, hidden dimensions are treated as continuous values and rounded post hoc, and $\sigma^2_{\mathbf{S}_l}$ is pre-calculated sparsely by population variance. Example C$^3$E solutions are provided in \ref{behavior}.

\begin{table*}[t]
  \centering
  \small
  % tighten up horizontal padding
  \setlength{\tabcolsep}{1mm}
  \begin{tabular*}{\textwidth}{@{\extracolsep{\fill}}lccccccccc}
    \toprule
    Model        & Cora     & Citeseer    & Pubmed     & AmzPhoto   & AmzComp    & Chameleon    & Squirrel   & ogb-arxiv & ogb-papers100M \\
    \midrule
    GCN          & $0.808_{\pm0.03}$ & $0.707_{\pm0.04}$ & $0.785_{\pm0.01}$ & $0.907_{\pm0.03}$ & $0.821_{\pm0.01}$ & $0.381_{\pm0.04}$ & $0.311_{\pm0.01}$ & $0.714_{\pm0.00}$ & $0.733_{\pm0.00}$    \\
    GCN$^\star$      & $\mathbf{\underline{0.837}}_{\pm0.07}$ & $\mathbf{0.723}_{\pm0.06}$ & $\mathbf{\underline{0.801}}_{\pm0.03}$ & $\mathbf{\underline{0.929}}_{\pm0.04}$ & $\mathbf{\underline{0.846}}_{\pm0.03}$ & $\mathbf{0.432}_{\pm0.09}$ & $\mathbf{\underline{0.346}}_{\pm0.08}$ & $\mathbf{\underline{0.729}}_{\pm0.00}$ & $\mathbf{\underline{0.760}}_{\pm0.00}$    \\
    \midrule
    APPNP        & $0.824_{\pm0.03}$ & $0.715_{\pm0.06}$ & $0.791_{\pm0.02}$ & $0.914_{\pm0.02}$ & $0.817_{\pm0.02}$ & $0.317_{\pm0.02}$ & $0.240_{\pm0.01}$ & $0.680_{\pm0.00}$ & $0.637_{\pm0.00}$    \\
    APPNP$^\star$    & $\mathbf{0.833}_{\pm0.05}$ & $\mathbf{0.724}_{\pm0.05}$ & $\mathbf{0.800}_{\pm0.02}$ & $\mathbf{\underline{0.926}}_{\pm0.03}$ & $\mathbf{\underline{0.832}}_{\pm0.03}$ & $\mathbf{\underline{0.366}}_{\pm0.08}$ & $\mathbf{\underline{0.282}}_{\pm0.05}$ & $\mathbf{\underline{0.704}}_{\pm0.00}$ & $\mathbf{\underline{0.681}}_{\pm0.00}$    \\
    \midrule
    GDC\textsubscript{HK}      & $0.826_{\pm0.02}$ & $0.718_{\pm0.03}$ & $0.792_{\pm0.02}$ & $0.920_{\pm0.02}$ & $0.832_{\pm0.03}$ & $0.335_{\pm0.01}$ & $0.262_{\pm0.02}$ & $0.679_{\pm0.00}$ & $0.667_{\pm0.00}$    \\
    GDC\textsubscript{HK}$^\star$  & $\mathbf{0.835}_{\pm0.04}$ & $\mathbf{0.726}_{\pm0.04}$ & $\mathbf{0.797}_{\pm0.04}$ & $\mathbf{0.928}_{\pm0.05}$ & $\mathbf{\underline{0.850}}_{\pm0.05}$ & $\mathbf{\underline{0.372}}_{\pm0.08}$ & $\mathbf{\underline{0.302}}_{\pm0.06}$ & $\mathbf{\underline{0.692}}_{\pm0.00}$ & $\mathbf{\underline{0.671}}_{\pm0.00}$  \\
    \midrule
    GDC\textsubscript{PPR}     & $0.824_{\pm0.02}$ & $0.720_{\pm0.02}$ & $0.789_{\pm0.01}$ & $0.910_{\pm0.02}$ & $0.830_{\pm0.03}$ & $0.330_{\pm0.02}$ & $0.264_{\pm0.01}$ & $0.677_{\pm0.00}$ & $0.650_{\pm0.00}$    \\
    GDC\textsubscript{PPR}$^\star$ & $\mathbf{\underline{0.837}}_{\pm0.04}$ & $\mathbf{0.725}_{\pm0.03}$ & $\mathbf{\underline{0.798}}_{\pm0.03}$ & $\mathbf{\underline{0.925}}_{\pm0.05}$ & $\mathbf{0.843}_{\pm0.06}$ & $\mathbf{\underline{0.377}}_{\pm0.05}$ & $\mathbf{\underline{0.309}}_{\pm0.06}$ & $\mathbf{\underline{0.695}}_{\pm0.00}$ & $\mathbf{\underline{0.688}}_{\pm0.00}$    \\
    \midrule
    SGC          & $0.783_{\pm0.01}$ & $0.700_{\pm0.02}$ & $0.753_{\pm0.01}$ & $0.869_{\pm0.02}$ & $0.808_{\pm0.01}$ & $0.287_{\pm0.01}$ & $0.231_{\pm0.02}$ & $0.696_{\pm0.00}$ & $0.660_{\pm0.00}$ \\
    SGC$^\star$      & ${\mathbf{\underline{0.827}}}_{\pm0.02}$ & ${\mathbf{\underline{0.719}}}_{\pm0.02}$ & ${\mathbf{\underline{0.790}}}_{\pm0.04}$ & ${\mathbf{\underline{0.915}}}_{\pm0.02}$ & ${\mathbf{\underline{0.827}}}_{\pm0.03}$ & ${\mathbf{\underline{0.334}}}_{\pm0.04}$ & ${\mathbf{\underline{0.269}}}_{\pm0.05}$ & $\mathbf{\underline{0.712}}_{\pm0.00}$ & $\mathbf{\underline{0.679}}_{\pm0.00}$   \\
    \midrule
    S\textsuperscript{2}GC     & $0.829_{\pm0.03}$ & $0.718_{\pm0.03}$ & $0.795_{\pm0.01}$ & $0.919_{\pm0.04}$ & $0.829_{\pm0.03}$ & $0.398_{\pm0.02}$ & $0.312_{\pm0.01}$ & $0.707_{\pm0.00}$ & $0.715_{\pm0.00}$    \\
    S\textsuperscript{2}GC$^\star$ & $\mathbf{\underline{0.841}}_{\pm0.04}$ & $\mathbf{0.724}_{\pm0.05}$ & $\mathbf{\underline{0.803}}_{\pm0.03}$ & $\mathbf{0.928}_{\pm0.05}$ & $\mathbf{\underline{0.847}}_{\pm0.04}$ & ${\mathbf{\underline{0.435}}}_{\pm0.06}$ & ${\mathbf{\underline{0.352}}}_{\pm0.06}$ & $\mathbf{\underline{0.726}}_{\pm0.00}$ & $\mathbf{\underline{0.753}}_{\pm0.00}$    \\
    \midrule
    JacobiConv       & $0.827_{\pm0.01}$ & $0.722_{\pm0.01}$ & $0.799_{\pm0.01}$ & $0.924_{\pm0.03}$ & $0.838_{\pm0.02}$ & $0.423_{\pm0.02}$ & $0.328_{\pm0.02}$ & $0.718_{\pm0.00}$ & $0.722_{\pm0.00}$    \\
    JacobiConv$^\star$   & $\mathbf{\underline{0.841}}_{\pm0.09}$ & $\mathbf{0.729}_{\pm0.05}$ & $\mathbf{\underline{0.807}}_{\pm0.02}$ & $\mathbf{0.928}_{\pm0.06}$ & $\mathbf{0.849}_{\pm0.04}$ & ${\mathbf{\underline{0.469}}}_{\pm0.06}$ & $\mathbf{0.351}_{\pm0.07}$ & $\mathbf{\underline{0.730}}_{\pm0.00}$ & $\mathbf{\underline{0.759}}_{\pm0.00}$    \\
    \midrule
    GPRGNN       & $0.821_{\pm0.01}$ & $0.692_{\pm0.01}$ & $0.792_{\pm0.02}$ & $0.917_{\pm0.02}$ & $0.824_{\pm0.01}$ & $0.348_{\pm0.02}$ & $0.243_{\pm0.02}$ & $0.711_{\pm0.00}$ & $0.654_{\pm0.00}$ \\
    GPRGNN$^\star$   & $\mathbf{\underline{0.843}}_{\pm0.08}$ & $\mathbf{\underline{0.728}}_{\pm0.07}$ & $\mathbf{\underline{0.809}}_{\pm0.04}$ & $\mathbf{\underline{0.931}}_{\pm0.08}$ & $\mathbf{\underline{0.847}}_{\pm0.05}$ & $\mathbf{0.379}_{\pm0.06}$ & ${\mathbf{\underline{0.288}}}_{\pm0.07}$ & $\mathbf{\underline{0.720}}_{\pm0.00}$ & $\mathbf{\underline{0.668}}_{\pm0.00}$    \\
    \midrule
    ChebNetII    & $0.822_{\pm0.01}$ & $0.696_{\pm0.01}$ & $0.791_{\pm0.01}$ & $0.908_{\pm0.03}$ & $0.815_{\pm0.03}$ & $0.430_{\pm0.04}$ & $0.336_{\pm0.01}$ & $0.720_{\pm0.00}$ & $0.670_{\pm0.00}$ \\
    ChebNetII$^\star$& $\mathbf{\underline{0.842}}_{\pm0.09}$ & ${\mathbf{\underline{0.727}}}_{\pm0.04}$ & ${\mathbf{\underline{0.807}}}_{\pm0.03}$ & $\mathbf{\underline{0.923}}_{\pm0.06}$ & ${\mathbf{\underline{0.848}}}_{\pm0.05}$ & $\mathbf{0.466}_{\pm0.07}$ & $\mathbf{0.357}_{\pm0.09}$ & $\mathbf{\underline{0.733}}_{\pm0.00}$ & $\mathbf{\underline{0.691}}_{\pm0.00}$    \\
    \bottomrule
  \end{tabular*}
  \caption{The semi-supervised node classification results (random splits and averaged over 10 runs) and node property prediction results (public splits and averaged over 10 runs). Starred models ($^\star$) are C$^3$E‐estimated baselines; \textbf{bold} fonts mark better average performance and \underline{underlines} mark statistical significance (t-test, $p<0.05$).}
  \label{perfcomp}
\end{table*}

\section{Experiments}
\label{sec7}
In this work, we conduct semi-supervised node classification with eight GNNs on seven public graphs: Cora \citep{sen2008collective}, Citeseer \citep{sen2008collective}, Pubmed~\citep{sen2008collective}, AmazonPhoto \citep{shchur2018pitfalls}, AmazonComputer \citep{shchur2018pitfalls}, Chameleon~\citep{rozemberczki2021multi}, and Squirrel~\citep{rozemberczki2021multi}. In addition, we perform node property prediction on two large-scale graphs: ogb-arxiv~\citep{hu2020open} and ogb-papers100M~\citep{hu2020open}. The hyperparameter configurations are provided in \ref{config}.

\subsection{Performance Evaluation}
\label{perf}
Figure~\ref{effex_analysis} and Table~\ref{perfcomp} illustrate that C$^3$E-estimated optimal models consistently outperform baselines using heuristic configurations across all scenarios. Figure~\ref{effex_analysis} visually confirms this: we see that many heuristic solutions (green dots) achieve scattered and suboptimal performance, whereas C$^3$E-estimated solutions reach optimal performance regions. This empirically validates that optimal performance is achieved when $\phi_0$ falls within bounds defined in Eq.~(\ref{lbcons}), and that simply scaling up parameters does not guarantee better results. Meanwhile, these results verify that while C$^3$E is formulated based on the principle of maximum entropy for pre-training estimation, C$^3$E-estimated models learn stable representations, in contrast to ill-behaved representations in naively configured models (see \ref{post-distri} for post-training analysis). 
Furthermore, unlike trial-and-error, which can take between 2.45 hours to 120 hours~\citep{cai2021rethinking}, our method generates solutions within seconds (3.7 seconds to 879.6 seconds) as shown in Table~\ref{time}.

\subsection{Representation Compression Ratio}
\label{ewsgnn}
As shown in Table~\ref{abas}, the representation compression ratio $\theta$ monotonically rises as the propagation depth $L$ increases. Increasing propagation depth consistently compresses information in the encoded representation, which aligns with previous empirical findings~\citep{saxe2019information,loukas2020hard,di2023over} and Corollary~\ref{theo2}. The C$^3$E-estimated baseline achieves optimal performance when $\theta = 0.010$ and $L = 4$. Conversely, plain baselines consistently degrade as $\theta$ increases beyond their minima ($\theta =0.558$ when $L=1$). These changes in model performance and representation compression ratio imply that over-squashing arises from cumulative information compression (increasing $\theta$). 
\begin{table}[t]
  \centering
  \begin{tabular}{@{}l|cccccc@{}}
    \toprule
    $L$ & $\overline w$ & $\theta$ & GCN & $\overline w$ & $\theta$ & GCN$^\star$ \\
    \midrule
    1 & 16 & 0.558 & \textbf{0.707} & 32765   & 0.000  & -              \\
    2 & 16 & 0.856 & 0.663          & 3960.89 & 0.004  & 0.716          \\
    3 & 16 & 1.155 & 0.624          & 3453.01 & 0.007  & 0.717          \\
    4 & 16 & 1.450 & 0.612          & 3203.64 & 0.010  & \textbf{0.723} \\
    5 & 16 & 1.750 & 0.605          & 2975.99 & 0.014  & 0.715          \\
    6 & 16 & 2.050 & 0.572          & 2808.91 & 0.017  & 0.713          \\
    7 & 16 & 2.347 & 0.550          & 2399.07 & 0.021  & 0.714          \\
    \bottomrule
  \end{tabular}
\caption{Comparison results between the baseline and C$^3$E-estimated baseline on Citeseer ($n=3312$, $m=3703$). Here, the empty cell denotes no valid solution.}
  \label{abas}
\end{table}
\begin{figure*}[tbhp]
    \centering
    \includegraphics[width=0.95\linewidth]{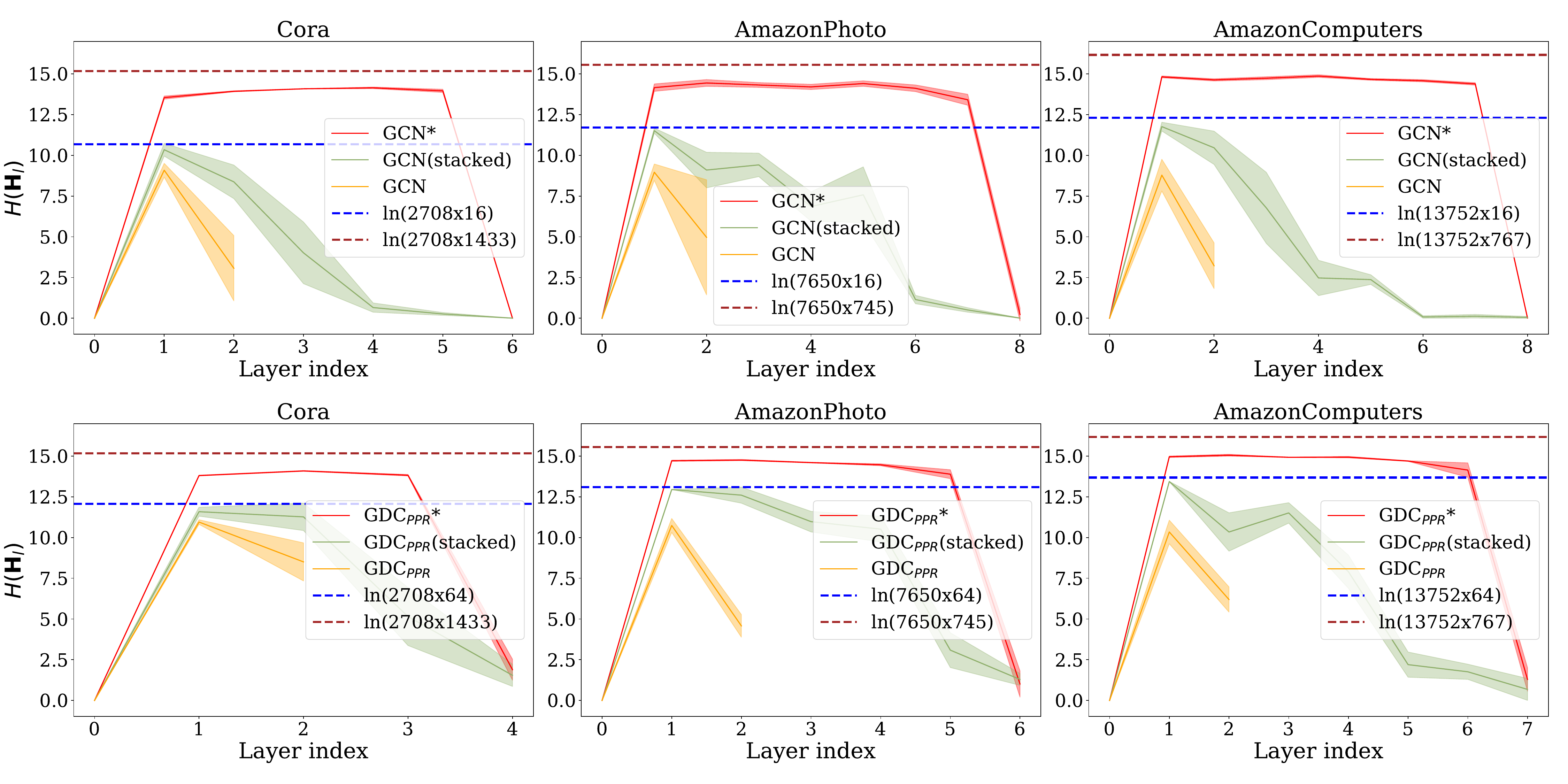}
    \label{sfigure1}
    \caption{C$^3$E-estimated models (red lines) avoid information loss by maintaining high $H(\mathbf{H}_l)$ across layers. In contrast, naively stacked baselines (green lines) suffer from over-squashing, with entropy collapsing to near-zero long before the final layer. Dotted lines indicate the maximum entropy for initial feature dimensions (brown) and fixed hidden dimensions (blue).}
    \label{entropy_analysis}
\end{figure*}
First, larger hidden dimensions reduce $\theta$, alleviating the information compression; however, overly small $\theta$ ($\overline{w}$ relatively larger) results in the encoded representation matrix becoming overly informative, and overestimates the latent distribution's complexity. In such cases, increasing propagation depth $L$ is not detrimental, facilitating compression of high-dimensional representations into appropriately parameterized lower-dimensional ones. In general, these results empirically verify Corollary~\ref{theo2} with further supportive visualizations provided in \ref{viztheta}.

\subsection{Entropy Transitions in Representation Matrix}
This section illustrates the layer-wise representation entropy transitions of original baselines, C$^3$E-estimated baselines, and naively stacked baselines. Based on previous analysis, over-squashing originates from cumulative growth in the information compression. Consequently, this should result in entropy reductions in encoded representation matrices. Figure~\ref{entropy_analysis} further visualizes the consequence of cumulative information compression on the network's information content. The entropy of naively stacked baselines (green lines) quickly collapses during propagation, demonstrating catastrophic information loss. This visual trend is the direct result of the runaway representation compression ratio (same in Table~\ref{abas}), where rising compression chokes off information flow. In other words, effective representation learning terminates early, causing encoded representation matrices to fail to preserve information about the latent distribution. In contrast, C$^3$E-estimated models maintain consistently high entropy representation learning, successfully preventing information loss and enabling effective representation learning during the information propagation. In addition, we can observe that the representation entropy of baselines using 16 or 64 as hidden dimensions is strictly bounded by their dimensionalities, as they do not exceed the blue dotted lines. This observation reiterates that the information conveyed in representation matrices is capped by their dimensions.

\section{Conclusion}
\label{conclusion}
This paper presents Channel Capacity Constrained Estimation (C$^3$E), an information-theoretic framework for estimating the hidden dimensions and depth for GNNs to mitigate over-squashing. Our analysis and results lead to three primary conclusions. First, through the lens of information theory, we show that over-squashing is a direct consequence of cumulative information compression within representation matrices. Second, ensuring hidden dimensions are sufficiently large can effectively mitigate such information compression. Third, the role of network depth is nuanced: for networks with a low representation compression ratio, deeper propagation is beneficial, helping concentrate high-dimensional signals to lower dimensions; for those with a high ratio, it aggravates information compression and leads to information loss. 

Despite the promising results, we acknowledge the limitations of this work. First, the framework is grounded on the principle of maximum entropy, which may overestimate what practical networks can achieve, but offers a useful theoretical ceiling before training. Second, though theoretical derivations are established with analytically tractable spectral GNNs, the results uncovered provide a critical basis for future extensions to more complex GNN architectures. The primary focus for future research is to generalize C$^3$E to spatial GNNs that depend on learned features for propagation, such as Graph Transformers. Moreover, we plan to tighten the entropy upper bounds with large-scale empirical results to more closely reflect practical trained GNNs.

\section{Acknowledgments}
This work was supported by UK Research and Innovation (UKRI) Engineering and Physical Sciences Research Council (EPSRC) Grant Number EP/Y028392/1: AI for Collective Intelligence (AI4CI).

\bigskip

%--- BIBLIOGRAPHY ---%
\bibliography{aaai2026}

\newpage
\onecolumn

\appendix
% Change numbering to letters
% Label sections as "Appendix A", "Appendix B", etc.
\renewcommand{\thesection}{Appendix~\Alph{section}}
% Subsections as "A.1", "A.2", etc. (no "Appendix")
\renewcommand{\thesubsection}{\Alph{section}.\arabic{subsection}}
% Make lemma numbers A.1, A.2, etc.
\renewcommand{\thetheorem}{\Alph{section}.\arabic{theorem}}
% Reset the theorem counter for Appendix A
\setcounter{theorem}{0}
\setcounter{section}{0} % restart numbering

% --- Force left-justified section headings ---
\makeatletter
\renewcommand{\section}{\@startsection{section}{1}{0pt}%
  {-2ex plus -0.5ex minus -.2ex}% Space before
  {1ex plus .2ex}% Space after
  {\normalfont\Large\bfseries\raggedright}} % <-- raggedright = left-justified
\makeatother

%--- APPENDICES ---%
% Appendix for AAAI-26: How Wide and How Deep? Mitigating Over-squashing of GNNs via Channel Capacity Constrained Estimation. You, Zheng, and Cartlidge

\section{Preliminary}
\label{prooflim}
\subsection{Justification of Matrix Entropy Definition}
\label{matenapp}
One might argue that the entry-wise definition for the matrix entropy in the preliminary can ignore any structural dependencies of entries. A fully structure-aware entropy would account for correlations among entries, precisely what the von Neumann entropy~\citep{von2018mathematical} does. For a density matrix $\rho\in\mathbb{C}^{N\times N}$, which must be Hermitian, positive semidefinite (PSD), and unit trace, it is defined as,
\begin{equation}
    H_{\text{v}}(\rho) = -\text{tr}(\rho\ln{(\rho))}.
\end{equation}
Here, $\text{tr}(\cdot)$ is the trace. The von Neumann entropy depends on the eigenvalue spectrum, thereby encoding global structure and inter‐entry dependencies~\cite{von2018mathematical}. Unfortunately, any real-valued finite-dimensional weight matrix or representation matrix $\mathbf{Z} \in \mathbb{R}^{\alpha \times \beta}$ in a neural network generally fails to satisfy the density‐matrix requirements. One could force it into the valid form by imposing the mapping,
\begin{equation}
  \rho = \frac{\mathbf{Z}^\top \mathbf{Z}}{\mathrm{tr}(\mathbf{Z}^\top \mathbf{Z})}.
  \label{eq:normalize_psd}
\end{equation}
However, this definition introduces three critical issues, 
\begin{itemize}
    \item It discards the original sign and relative scale of the entries.
    \item It couples all entries through the global trace normalization, so local (entry-wise) uncertainties are obliterated. 
    \item The resulting density matrix $\rho$ is always square, which discards the original dimensionality of a non-square matrix $\mathbf{Z} \in \mathbb{R}^{\alpha \times \beta}$ and imposes an unnatural structural constraint.
\end{itemize}
As a result, von Neumann entropy of the normalized $\rho$ bears little relation to the actual uncertainty or interpretability of $\mathbf{Z}$~\citep{cover1999elements,shannon1948mathematical,jaynes1957information}. 

Conversely, entry-wise Shannon entropy imposes only the minimal assumption that each observed value $z_i$ occurs with nonnegative probability summing to one. Viewing the entries of $\mathbf{Z}$ as independent draws from a random variable $Z\sim p$ (thus avoiding any intractable covariance structure), one can write,
\[\boxed{
H(Z) \;=\;
\begin{cases}
-\displaystyle\int p(z)\ln p(z)\,dz, 
& \text{(continuous support)},\\[1ex]
-\displaystyle\sum_{i=1}^{|\mathrm{Supp}(Z)|}P(Z=z_i)\ln P(Z=z_i),
& \text{(discrete support)}.
\end{cases}}
\]
This definition enjoys three practical advantages:
\begin{itemize}
  \item It applies to any real matrix, regardless of sign, symmetry, or boundedness.
  \item It can be estimated efficiently via histogramming or kernel‐density estimation.
  \item It directly quantifies the average information content of the entries.
\end{itemize}
Although viewing the entries of $\mathbf{Z}$ as independent draws from $Z \sim p$ disregards higher-order correlations, in many neural-network settings (including GNNs that are explicitly designed to induce correlations between node representations based on graph topology), entries are only weakly coupled by weight-sharing or convolutional locality~\citep{schoenholz2016deep,jacot2018neural,wang2023theoretical}, thus ignoring high-order correlations is often reasonable.

One could, in principle, estimate higher‐order correlations via joint entropy, but that rapidly becomes computationally intractable (e.g., for $N$ entries in $\mathbf{Z}$ leads to $\mathcal{O}[(\alpha \beta)^N]$), and still struggles to resolve the sign and scale issues mentioned previously. Therefore, our definition remains the most practical and assumption-light proxy for matrix uncertainty when no prior guarantees (e.g.\ PSD, Hermiticity, unit trace) are available~\citep{shannon1948mathematical,jaynes1957information}.

\subsection{Proof of Entropy Upper Bounds}
\label{proofenup}
\begin{lemma}
Among all real-valued continuous random variables $Z$ given mean $\mu_{Z}$ and variance $\sigma^2_{Z}$, the Gaussian distribution $\mathcal{N}(\mu_{Z}, \sigma^2_{Z})$ achieves the maximum entropy, which is
\begin{align}
\boxed{H(Z) \leq \frac{1}{2} \ln{\left( 2 \pi e \sigma^2_{Z} \right)}.}
\end{align}
\label{prooflemma1}
\end{lemma}
\textbf{Proof.} Let $p(z)$ be the probability density function (PDF) of $Z$, and let $p_G(z)$ be the PDF of a Gaussian $Z_G \sim \mathcal{N}(\mu_{Z}, \sigma^2_{Z})$. The difference in their entropies can be related to the Kullback-Leibler (KL) divergence. A full proof based on the non-negativity of KL-divergence, $D_{KL}(p || p_G) \geq 0$, can be found in standard texts~\citep{cover1999elements}, which confirms that $H(Z) \leq H(Z_G)$.

The entropy of the Gaussian distribution $H(Z_G)$ is calculated as:
\begin{align*}
H(Z_G) &= -\int_{-\infty}^{\infty} p_G(z) \ln{p_G(z)} \, dz \\
&= -\int_{-\infty}^{\infty} p_G(z) \left( -\ln{\sqrt{2 \pi \sigma^2_{Z}}} - \frac{(z - \mu_{Z})^2}{2\sigma^2_{Z}} \right) dz \\
&= \ln{\sqrt{2 \pi \sigma^2_{Z}}} \int p_G(z) \, dz + \frac{1}{2\sigma^2_{Z}} \int p_G(z) (z - \mu_{Z})^2 \, dz \\
&= \ln{\sqrt{2 \pi \sigma^2_{Z}}} \cdot 1 + \frac{1}{2\sigma^2_{Z}} \cdot \sigma^2_{Z} \\
&= \frac{1}{2} \ln{(2 \pi \sigma^2_{Z})} + \frac{1}{2}\\
&= \frac{1}{2} \ln{\left( 2 \pi \mathrm{e} \sigma^2_{Z} \right) }.
\end{align*}
Since $H(Z) \leq H(Z_G)$, the lemma holds.

\begin{lemma}
For a discrete random variable $Z$ with a finite support set $\mathrm{Supp}(Z)$, the entropy is bounded by $H(Z) \leq \ln{(|\mathrm{Supp}(Z)|)}$. For a matrix with $\alpha\beta$ entries, the support size is at most $\alpha\beta$, thus:
\begin{equation}
 \boxed{H(Z) \leq \ln{(|\mathrm{Supp}(Z)|)} \leq \ln{(\alpha\beta)}.}
 \label{appena2}
\end{equation}
\label{prooflemma2}
\end{lemma}
\textbf{Proof.} The entropy of a discrete random variable $Z$ is $H(Z) = -\sum_{i} P(z_i) \ln{P(z_i)}$. From Jensen's inequality, since $\ln{(x)}$ is a concave function, the entropy is maximized when the probability distribution is uniform over its support set~\citep{cover1999elements}. 

Let the support size be $k=|\mathrm{Supp}(Z)|$. For the uniform distribution, $P(Z = z_i) = \frac{1}{k}$ for all $z_i \in \mathrm{Supp}(Z)$. The entropy is:
\begin{align*}
 H(Z) &= -\sum_{i=1}^{k} \frac{1}{k} \ln{\left( \frac{1}{k} \right)} = -k \left( \frac{1}{k} \ln{\left( \frac{1}{k} \right)} \right) = -\ln{\left( \frac{1}{k} \right)} = \ln{(k)}.
\end{align*}
Since the number of distinct values (the support size) in a matrix of size $\alpha \times \beta$ cannot exceed the number of entries, we have $k \leq \alpha\beta$. Therefore, $H(Z) \leq \ln{(k)} \leq \ln{(\alpha\beta)}$.
One should know that in practice, we only have a finite number of samples (entries), which effectively discretizes the unknown distribution. Although such discretization introduces approximation errors, the discrete form converges to the continuous form as the number of samples increases and the discretization becomes finer.

% Next appendix: reset counter again
\setcounter{theorem}{0}
\section{Proof of Theorem 1}
\label{proof4.1}
\subsection{Gaussian and Independence across Matrices}
\label{inde}

Before updating weights in GNNs, we have no information about the joint behavior of the entries in either the representation matrix or the learnable weight matrix of the network, which provides information about the unknown distribution. In particular, we only know that the mean and variance of these matrices as random variables ($Z_1,\cdots,Z_i$) exist, and we do not know any correlations between different matrices. Under such a situation, we need an observable standpoint to perform the estimation of the network.

Fortunately, Jaynes’s principle of maximum entropy \citep{jaynes1957information,jaynes2003probability} offers the most unbiased and representative view about neural networks in this scenario. It states that the least‐informative or most unbiased prior satisfying a given set of moment constraints is the one that maximizes entropy subject to those constraints. Here, our constraints are exactly:
\[
\forall i:\;
\mathbb{E}[Z_i] = \mu_{Z_i},\;
\mathbb{E}\bigl[(Z_{i} - \mu_{Z_i})^2\bigr] = \sigma^2_{Z_i},
\]
and no constraints on any covariances $\mathbb{E}[(Z_i -  \mu_{Z_i})(Z_j - \mu_{Z_j})]$ for $i\neq j$.

\textbf{Gaussian}. Lemma A.1 tells us that, among any real‐valued variable with specific mean and variance, the Gaussian has the largest entropy. Hence each marginal $Z_i$ must be Gaussian:
    \[
      p_i(Z_i) = \mathcal{N}(Z_i; \mu_{Z_i},\sigma^2_{Z_i}).
    \]

\textbf{Independence}. Since we have imposed no constraints relating different $Z_i$, the maximum‐entropy joint density subject to the individual‐marginal constraints is simply the product of the marginals:
    \[
      p(Z_1,\dots,Z_N)
      \;=\;\prod_{i=1}^N p_i(Z_i).
    \]
    Any nonzero covariance would introduce extra structure (i.e.\ impose additional constraints) and therefore reduce entropy. Putting these facts together, the maximum‐entropy prior over any weight (or representation) matrix is an independent Gaussian:
$$ \boxed{ p(Z_1,\dots,Z_N) \;=\; \prod_{i=1}^n \mathcal{N}\bigl(Z_i;\,\mu_{Z_i},\sigma^2_{Z_i}\bigr),}$$ which encodes maximum entropy and independence among matrices.

\begin{lemma}
    Let $\{Z_1, Z_2, \ldots, Z_o\}$ be a set of independent random variables. Then, the expectation and variance of their sum and product satisfy the following properties:
    \begin{align}
        \mathbb{E}\left(\sum_{i=1}^{o} Z_i\right) &= \sum_{i=1}^{o} \mathbb{E}(Z_i), \label{eq:expectation_sum} \\
        \sigma^2\left(\sum_{i=1}^{o} Z_i\right) &= \sum_{i=1}^{o} \sigma^2(Z_i), \label{eq:variance_sum} \\
        \mathbb{E}\left(\prod_{i=1}^{o} Z_i\right) &= \prod_{i=1}^{o} \mathbb{E}(Z_i), \label{eq:expectation_product} \\
        \sigma^2\left(Z_i Z_j\right) &= \sigma^2(Z_i)\sigma^2(Z_j) + \sigma^2(Z_i)\mathbb{E}^2(Z_j) + \mathbb{E}^2(Z_i)\sigma^2(Z_j), \label{eq:variance_product}
    \end{align}
    for all $i, j = 1, 2, \ldots $ and $i \neq j$.
\end{lemma}

\textbf{Proof.} {Expectation of the Sum:}
    \[
        \mathbb{E}\left(\sum_{i=1}^{o} Z_i\right) = \sum_{i=1}^{o} \mathbb{E}(Z_i).
    \]
    
    {Variance of the Sum:}
    Since the random variables are independent,
    \[
        \sigma^2\left(\sum_{i=1}^{o} Z_i\right) = \sum_{i=1}^{o} \sigma^2(Z_i).
    \]
    
    {Expectation of the Product:}
    For independent random variables,
    \[
        \mathbb{E}\left(\prod_{i=1}^{o} Z_i\right) = \prod_{i=1}^{o} \mathbb{E}(Z_i).
    \]
    
    {Variance of the Product:}
    For two independent random variables $Z_i$ and $Z_j$:
    \[
        \sigma^2(Z_i Z_j) = \mathbb{E}[(Z_i Z_j)^2] - \mathbb{E}[Z_i Z_j]^2.
    \]
    Since $Z_i$ and $Z_j$ are independent,
    \[
        \mathbb{E}[(Z_i Z_j)^2] = \mathbb{E}[Z_i^2] \mathbb{E}[Z_j^2] = \sigma^2(Z_i) + \mathbb{E}^2(Z_i), \; \mathbb{E}[Z_i Z_j] = \mathbb{E}[Z_i] \mathbb{E}[Z_j].
    \]
    Therefore,
    \begin{align}
                \sigma^2(Z_i Z_j) & =  (\sigma^2(Z_i) + \mathbb{E}^2(Z_i))(\sigma^2(Z_j) + \mathbb{E}^2(Z_j)) - \mathbb{E}^2(Z_i) \mathbb{E}^2(Z_j)\notag\\
                &= \sigma^2(Z_i)\sigma^2(Z_j) + \sigma^2(Z_i)\mathbb{E}^2(Z_j) + \mathbb{E}^2(Z_i)\sigma^2(Z_j).\notag
    \end{align}

\subsection{Invariance of Maximum Entropy under Element-wise Activation}
\label{act-invar}
Recall that for any finite support of size $N$, the highest achievable entropy is $\ln{(N)}$ according to Lemma~\ref{prooflemma2}. Now consider the pre-activation matrix $\mathbf{Q}\in \mathbb{R}^{\alpha\times \beta}$. Let $\mathrm{Supp}(Q)$ denote the finite set of possible scalar values that the entries $\mathbf{Q}^{i,j}$ can take. Let $\Delta(\cdot)$ be any element-wise nonlinearity (e.g., ReLU, sigmoid, tanh). Define the post-activation matrix,
\[\mathbf{Q}' = \Delta(\mathbf{Q}).\]
\textbf{No increase in support size}. Since $\Delta(\cdot)$ acts independently on $\mathbf{Q}^{i,j}$ to produce $(\mathbf{Q'})^{i,j}$, the support of $\mathbf{Q}^{'}$ is,
\[\mathrm{Supp}({Q}^{'}) = \{\Delta(\mathbf{Q}^{i,j}): \mathbf{Q}^{i,j}\in \mathrm{Supp}({Q}) \}.\] In particular,
\[|\mathrm{Supp}({Q}^{'})| \leq |\mathrm{Supp}({Q})|.\]
Since an element-wise mapping cannot produce more distinct output values than the number of distinct input values it receives. Then, together with Lemma~\ref{prooflemma2} we have,
$$\boxed{H(\mathbf{Q}')\leq H_{\max}(\mathbf{Q}).}$$

\subsection{Proof of the Entropy of Encoded Representation}
\label{theo4.1}
We treat $nw_l$ entries in $\mathbf{H}_l$ as outcomes from a scalar random variable as stated in Appendix~A.1. Unless noted otherwise, we report entropy per entry; multiplying by $nw_l$ yields the joint entropy of the whole matrix. This per-channel-use convention matches the classic definition of channel capacity in information theory~\citep{shannon1948mathematical}.

\textbf{Proof}. As established in Appendix~\ref{inde}, we approach the estimation from the standpoint of maximum entropy, assuming that before training, the network's components are in a state of maximal uncertainty subject to their basic constraints. This implies that the distributions of entries in the weight and representation matrices are independent and Gaussian. Furthermore, as shown in Appendix~\ref{act-invar}, element-wise activation functions do not increase the maximum entropy of a representation. We can therefore analyze the information flow as follows:
\begin{align}    
H(\mathbf{H}_{l}) & = H_{\max}(\Delta(\mathbf{S}_l \mathbf{H}_{l-1} \mathbf{W}_l)) \leq H_{\max}(\mathbf{S}_l \mathbf{H}_{l-1} \mathbf{W}_l) \label{eq:H_pre_activation}
\end{align}
We proceed by calculating the mean and variance of the representation at each layer $l$. Based on Appendix~B.1, Appendix~B.2, and Lemma~B.1. We perform the following derivations. 

\textbf{Mean of representation.} We assume the initial features $\mathbf{H}_0$ are normalized to be zero-mean and unit variance, i.e., $\mathbb{E}[\mathbf{H}_0^{i,j}] = 0, \; \sigma^2_{H_0}=1$. The mean of the representation at any layer $l \ge 1$ remains zero, as shown by induction:
\[ \mathbb{E}[\mathbf{H}_{l-1}] = \mathbf{0} \implies \mathbb{E}[\mathbf{S}_l \mathbf{H}_{l-1}] = \mathbf{S}_l \mathbb{E}[\mathbf{H}_{l-1}] = \mathbf{0}. \]
Since the weights $\mathbf{W}_l$ are independent of the input $\mathbf{H}_{l-1}$:
\[ \mathbb{E}[\mathbf{H}_l] = \mathbb{E}[(\mathbf{S}_l\mathbf{H}_{l-1})\mathbf{W}_l] = \mathbb{E}[\mathbf{S}_l\mathbf{H}_{l-1}] \mathbb{E}[\mathbf{W}_l] = \mathbf{0} \cdot \mathbb{E}[\mathbf{W}_l] = \mathbf{0}. \]
Thus, the mean of representations at all layers is zero-mean.

\textbf{Variance of representation.} We derive the recursive formula for the variance of the representation entries, $\sigma^2_{H_l}$. First, we compute the variance of an entry in the pre-transform matrix $\mathbf{U}_l = \mathbf{S}_l \mathbf{H}_{l-1}$, which is given by,
\begin{align*}
\sigma^2_{U_l} = \sigma^2\left(\mathbf{U}_l^{i,j}\right) = \sigma^2\left( \sum_{q=1}^{n} \mathbf{S}_l^{i,q} \mathbf{H}_{l-1}^{q,j} \right) &= \sum_{q=1}^{n} (\mathbf{S}_l^{i,q})^2 \sigma^2(\mathbf{H}_{l-1}^{q,j}) \\
&= \sigma^2_{H_{l-1}} \sum_{q=1}^{n} (\mathbf{S}_l^{i,q})^2 \approx \sigma^2_{H_{l-1}} \cdot (n \sigma^2_{\mathbf S_l}).
\end{align*}
Now, we use this result to compute the variance of an entry in $\mathbf{H}_l$:
\begin{align}
\sigma^2_{H_l} &= \sigma^2\left(\sum_{s=1}^{w_{l-1}}\mathbf{U}_l^{i,s}\mathbf{W}_l^{s,j}\right) = \sum_{s=1}^{w_{l-1}} \sigma^2(\mathbf{U}_l^{i,s}\mathbf{W}_l^{s,j}) \nonumber \\
&= \sum_{s=1}^{w_{l-1}} \left( \sigma^2_{U_l}(\sigma^2_{w_l} + \mu_{w_l}^2) + \mathbb{E}^2(\mathbf{U}_l^{i,s})\sigma^2_{w_l} \right) \nonumber \\
&= \sum_{s=1}^{w_{l-1}} \left( \sigma^2_{U_l}(\sigma^2_{w_l} + \mu_{w_l}^2) + 0 \cdot \sigma^2_{w_l} \right)  \nonumber \\
&= w_{l-1} \cdot \sigma^2_{U_l} \cdot (\sigma^2_{w_l} + \mu_{w_l}^2) \nonumber \\
&= w_{l-1} \cdot (n \sigma^2_{S_l} \sigma^2_{H_{l-1}}) \cdot (\sigma^2_{w_l} + \mu_{w_l}^2) \nonumber \\
&= n\,w_{l-1}\,(\sigma^2_{w_l}+\mu_{w_l}^2)\,\sigma^2_{S_l}\,\sigma^2_{H_{l-1}}.
\label{eq:var_recursion}
\end{align}
By iterating this recurrence from layer $1$ to $L$, with initial variance $\sigma^2_{H_0}=1$, we get the final variance:
\[
\sigma^2_{H_L} = \prod_{l=1}^{L} \Bigl[n\,w_{l-1}\,(\sigma^2_{w_l}+\mu_{w_l}^2)\,\sigma^2_{S_l}\Bigr].
\]

\textbf{Final representation entropy formula.} The maximum entropy of the encoded representation $\mathbf{H}_L$ is given by the entropy of a Gaussian with variance $\sigma^2_{H_L}$:
\begin{align}
 \boxed{H(\mathbf{H}_L) = \frac{1}{2}\ln{(2\pi \mathrm{e}\sigma^2_{H_L})} =  \frac{1}{2}\ln{\left(2\pi \mathrm{e} \prod_{l=1}^{L}nw_{l-1}(\sigma^2_{{w}_l}+\mu_{{w}_l}^2)\sigma^2_{{S}_{l}}\right)}.}
\end{align}
 Consequently, the channel capacity formula is expressed as:
\begin{equation}
 \boxed{\phi = \max{H(\mathbf{H}_L)} =  \max{\left[\frac{1}{2}\ln{\left(2\pi \mathrm{e} \prod_{l=1}^{L}nw_{l-1}(\sigma^2_{{w}_l}+\mu_{{w}_l}^2)\sigma^2_{{S}_{l}}\right)}\right].}}
\end{equation}

To arrive at the final simplified form, we assume a weight initialization scheme that preserves signal variance. This implies that for each layer $l$, the expected squared weight is one, i.e., $\mathbb{E}[(\mathbf{W}_l^{i,j})^2] = \sigma^2_{w_l}+\mu_{w_l}^2 = 1$. This is a common objective in variance-aware initializations like those of Xavier or Kaiming. Consequently, the simplified form of the channel capacity is expressed as:
\begin{equation}
 \boxed{\phi = \max{H(\mathbf{H}_L)} = \max{\left[\frac{1}{2}\ln{(2\pi \mathrm{e})} + \frac{1}{2}\sum_{l=1}^{L}\ln{(nw_{l-1}\sigma^2_{S_{l}})}\right]}.}
\end{equation}
This completes the proof of Theorem 1.

% Next appendix: reset counter again
\setcounter{theorem}{0}
\section{Proof of Corollary 2}
\label{col2}
\subsection{Derivation of the Threshold $\bar{w}^*$}
We begin with the definitions for channel capacity $\phi$ and geometric mean of hidden dimensions $\bar{w}$:
$$ \phi = \frac{1}{2}\ln{(2\pi \mathrm{e})} + \frac{1}{2}\sum_{l=1}^{L}\ln{(nw_{l-1}\sigma^2_{S_{l}})},\; \bar{w} = \left(\prod_{l=1}^{L} w_l\right)^{\frac{1}{L}} $$
The key is to express the summation term $\sum_{l=1}^{L}\ln(w_{l-1})$ as a function of $\bar{w}$.
\begin{align*}
\sum_{l=1}^{L}\ln(w_{l-1}) &= \ln(w_0) + \ln(w_1) + \dots + \ln(w_{L-1}) \nonumber \\
&= \ln(w_0) + \left(\sum_{l=1}^{L}\ln(w_l)\right) - \ln(w_L) \nonumber \\
&= \ln(m) + L\ln(\bar{w}) - \ln(w_L)
\end{align*}
Substituting this back into the definition for $\phi$ gives $\phi$ as a function of $\bar{w}$. To find the maximum of $\theta(\bar{w}) = \phi(\bar{w})/\bar{w}$, we take the partial derivative with respect to $\bar{w}$ and set it to zero.
\begin{align*}
\frac{\partial\theta}{\partial\bar{w}} &= \frac{\partial}{\partial\bar{w}}\left( \frac{\phi(\bar{w})}{\bar{w}} \right) = 0\notag
\end{align*}
Solving for $\bar{w}$ yields the optimal value $\bar{w}^*$. After substitution and algebraic manipulation, we arrive at:
\begin{align*}
\ln(\bar{w}^*) &= 1 - \frac{A}{B} \\
\text{where } A &= \frac{1}{2}\left( \ln(2\pi e) + \sum_{l=1}^{L}\ln(n\sigma^2_{S_{l}}) + \ln(m) - \ln(w_L) \right) \quad \text{and} \quad B = \frac{L}{2} \nonumber
\end{align*}
This gives the final expression:
$$\boxed{ \bar{w}^* = \mathrm{e}^{\left[1 - \frac{\ln(2\pi e) + \ln(m) - \ln(w_L) + \sum_{l=1}^{L}\ln(n\sigma^2_{S_{l}})}{L}\right]}} $$

To find the limit as the depth $L$ approaches infinity, we evaluate the limit of the exponent's fractional term:
$$\lim_{L \to \infty} \frac{\ln(2\pi e) + \ln(m) - \ln(w_L) + \sum_{l=1}^{L}\ln(n\sigma_{S_l}^2)}{L}$$
We can break this into the sum of the limits of each term:
$$
\lim_{L \to \infty} \left( \frac{\ln(2\pi e)}{L} \right) + \lim_{L \to \infty} \left( \frac{\ln(m)}{L} \right) - \lim_{L \to \infty} \left( \frac{\ln(w_L)}{L} \right) + \lim_{L \to \infty} \left( \frac{\sum_{l=1}^{L}\ln(n\sigma_{S_l}^2)}{L} \right)
$$Evaluating each term individually:$$
\lim_{L \to \infty} \frac{\ln(2\pi e)}{L} = 0
$$
$$\lim_{L \to \infty} \frac{\ln(m)}{L} = 0$$
$$\lim_{L \to \infty} \frac{\ln(w_L)}{L} = 0$$
The final term is the definition of the average log propagation variance, $\overline{K}$ [1, 1]:
$$\lim_{L \to \infty} \frac{\sum_{l=1}^{L}\ln(n\sigma_{S_l}^2)}{L} = \mathbb{E}_{l}[\ln{(n\sigma^2_{S_l})}] = \overline{K}$$
Combining these results, the limit of the fractional term is simply $\overline{K}$. Substituting this back into the original expression gives the final result stated in Corollary 2:
$$\boxed{\lim_{L \to \infty} \overline{w}^{*} = e^{1 - \overline{K}}}$$

\subsection{Derivation of the Effect of Depth $L$ on $\theta$}
To analyze the trend of $\theta$ with respect to depth $L$ for a fixed average width $\bar{w}$, we first approximate $\phi$ as a function of $L$. We assume $w_l \approx \bar{w}$ and that the average log-propagation variance is $\bar{K} = \mathbb{E}_l[\ln(n\sigma^2_{S_l})]$.
\begin{align*}
\phi(L) &\approx \frac{1}{2}\ln(2\pi e) + \frac{1}{2}\left( \sum_{l=1}^{L}\bar{K} + \sum_{l=1}^{L}\ln(w_{l-1}) \right) \\
&\approx \frac{1}{2}\ln(2\pi e) + \frac{1}{2}\left( L\bar{K} + \ln(m) + (L-1)\ln(\bar{w}) \right) \\
&\approx \left( \frac{\ln(2\pi e) + \ln(m) - \ln(\bar{w})}{2} \right) + \frac{L}{2}(\bar{K} + \ln(\bar{w}))
\end{align*}
The compression ratio is $\theta(L) = \phi(L)/\bar{w}$. To find the trend, we analyze the sign of the partial derivative with respect to $L$, treating $L$ as a continuous variable for trend analysis.
$$ \boxed{\frac{\partial\theta}{\partial L} \approx \frac{\partial}{\partial L} \left[ \frac{1}{\bar{w}} \left( \text{constant} + \frac{L}{2}(\bar{K} + \ln(\bar{w})) \right) \right] = \frac{1}{2\bar{w}}\left(\bar{K} + \ln(\bar{w})\right)} $$
Since $2\bar{w} > 0$, the sign of the derivative is determined by the sign of $(\bar{K} + \ln(\bar{w}))$. This yields two regimes:
\begin{itemize}
    \item \textbf{Wide Regime:} If $\ln(\bar{w}) > -\bar{K}$, then $\partial\theta/\partial L > 0$, and $\theta$ increases with depth $L$.
    \item \textbf{Narrow Regime:} If $\ln(\bar{w}) \le -\bar{K}$, then $\partial\theta/\partial L < 0$, and $\theta$ decreases with depth $L$.
\end{itemize}
This completes the derivation of Corollary 2.

% Next appendix: reset counter again
\setcounter{theorem}{0}
\section{Justification of Effective Channel Capacity $\phi_0$}
\label{theo6.1}
We begin by noting a practical constraint on hidden representations in real-world GNNs: for classification or property prediction tasks, each layer must have a width of at least two ($w_{l},w_{l-1}\ge2$), which immediately implies a simple but useful inequality,
\begin{equation}
    1\ge \frac{w_{l}+w_{l-1}}{w_{l}w_{l-1}}.
    \label{begeq}
\end{equation}

\subsection{Why effective channel capacity?}
The theoretical channel capacity $\phi$ introduced in Theorem~1 quantifies the maximum total entropy a GNN could propagate through $L$ layers in the absence of any architectural constraints, effectively assuming infinite width and unrestricted information flow. However, real networks operate under strict structural bottlenecks, particularly finite layer widths and sparse inter-layer connectivity. These constraints fundamentally limit how much information can be transmitted between layers.

To account for such limitations, we introduce the notion of an effective channel capacity $\phi_0$, which quantifies the actual information-carrying capability of a finite-width GNN architecture.

First, we start with the entropy telescoping identity,
\begin{align}
H(\mathbf H_L)-H(\mathbf H_0)
   &=\sum_{l=1}^{L}
     \frac{H(\mathbf H_l)-H(\mathbf H_{l-1})}{H(\mathbf H_l)}
     \cdot H(\mathbf H_l),\notag\\
     &\ge\sum_{l=1}^{L}
     \frac{H(\mathbf H_l)-H(\mathbf H_{l-1})}{H(\mathbf H_l)}
     \cdot I(\mathbf H_l;\mathbf H_{l-1}),\notag
\end{align}
where the inequality follows from a basic property of information theory: for any pair of random variables, their mutual information is bounded above by the entropy of either variable, i.e., $H(\mathbf H_l) \ge I(\mathbf H_l;\mathbf H_{l-1})$. This leads us to define the effective channel capacity $\phi_0$, 
\begin{equation}
\boxed{\phi_0 \triangleq \sum_{l=1}^{L}
\underbrace{\frac{H(\mathbf H_l) - H(\mathbf H_{l-1})}{H(\mathbf H_l)}}_{\text{Information Retention Ratio}}
\cdot
\underbrace{I(\mathbf H_l;\mathbf H{l-1})}_{\text{Mutual Information}},}
\label{eq:MI_lower}
\end{equation}
which provides a guaranteed lower bound on the network’s total entropy gain. Unlike the idealized $\phi$, this measure explicitly incorporates architectural constraints via the layer-wise mutual information, capturing how much information can pass through each finite-width layer. We now derive closed-form expressions for the two quantities in Eq.~\eqref{eq:MI_lower}.

\subsection{Layer-wise Mutual Information under Maximum Entropy}
\label{sec:MI-derivation}
To quantify how much information is preserved between adjacent GNN layers, we analyze the mutual information between two consecutive representations under a maximum-entropy distributional model. This analysis reflects the informational coupling achievable given architectural constraints such as finite layer width.

Let $s = w_{l-1}$ and $q = w_{l}$ denote the widths of two adjacent layers. Consider a pair of aligned activation vectors $(\mathbf{x}, \mathbf{y}) \in \mathbb{R}^{s+q}$, with $\mathbf{x}$ representing a row from layer $l-1$ and $\mathbf{y}$ from layer $l$. From a structural perspective, the transformation from $\mathbf{x}$ to $\mathbf{y}$ is governed by the weight matrix $\mathbf{W}_l \in \mathbb{R}^{q \times s}$ with element-wise activations. This layer implements a fully connected bipartite mapping, with each of the $sq$ directed edges representing a learnable parameter. Topologically, this corresponds to a complete bipartite graph $\mathcal{K}_{s,q}$~\citep{shannon1948mathematical}. We refer to these connections collectively as edges, and the variances they induce as edge energies. Based on Eq.~\eqref{begeq}, we define,
\begin{equation}
r :=\frac{s+q}{s\,q},\; 0<r\le 1,
\label{eq:r-def-new}
\end{equation}
which measures how narrow the bipartite interface is: the larger $r$, the stricter the structural bottleneck. We model the joint distribution of adjacent representations using a maximum-entropy Gaussian with the following properties:
\begin{itemize}
    \item The marginals over $\mathbf{x}$ and $\mathbf{y}$ are isotropic, as established in Appendix~B.1.
    \item The dependence between $\mathbf{x}$ and $\mathbf{y}$ is captured via a single shared latent factor, encoded as a rank-1 correlation structure.
\end{itemize}
This gives rise to the following joint covariance matrix:
\begin{equation}
\Sigma(\alpha)
=\sigma^{2}
  \begin{pmatrix}
  (1-r^{2})\mathbf{I}_{s} & \alpha\mathbf 1_s\mathbf 1_q^{\!\top}\\
  \alpha\mathbf 1_q\mathbf 1_s^{\!\top} & (1-r^{2})\mathbf{I}_{q}
  \end{pmatrix},
\;
0\le\alpha<\frac{1-r^{2}}{\sqrt{s\,q}}.
\label{eq:Sigma}
\end{equation}
where $\sigma^2$ is the marginal variance identical across coordinates and equal $(1-r^2)\sigma^2$, and $\alpha$ controls the strength of inter-layer coupling.

Given these, the mutual information between $\mathbf{x}$ and $\mathbf{y}$ can be computed analytically. For block matrices of the form in Eq.~\eqref{eq:Sigma}, canonical correlation analysis yields~\citep{cover1999elements}:
\begin{equation}
I_{\alpha}(\mathbf y;\mathbf x)
  = -\frac12\ln\bigl(1-\frac{\alpha^{2}sq}{(1-r^2)^2}\bigr).
\label{eq:MI-alpha}
\end{equation}
This expression reflects the extent of shared variability between the two layers, which is governed entirely by $\varepsilon:= \frac{\alpha^2sq}{(1-r^2)^2}$, capturing the coupling energy. As $\alpha$ increases, the strength of dependence grows, thereby increasing mutual information. Hence, maximizing mutual information between adjacent layers is equivalent to maximizing $\alpha^2$ under structural and energetic constraints, as we explore in the following subsections.

\subsubsection{PSD Constraint on Coupling Strength}
To ensure the joint Gaussian is well-defined, the covariance matrix $\Sigma(\alpha)$ must be PSD. Working with the dimensionless form
\[
\mathbf M
:=\frac{\Sigma(\alpha)}{\sigma^{2}}
=
\begin{pmatrix}
(1-r^{2})\mathbf{I}_{s} & \alpha\mathbf 1_{s}\mathbf 1_{q}^{\!\top}\\
\alpha\mathbf 1_{q}\mathbf 1_{s}^{\!\top} & (1-r^{2})\mathbf{I}_{q}
\end{pmatrix}.
\]
Note that $(1-r^2)\mathbf{I}_{s}$ is PD for $r<1$ (and PSD at $r=1$). Applying the Schur-complement criterion~\citep{fuzhen2005schur} with respect to this block yields the condition,
\[
(1-r^{2})\mathbf{I}_{q}-\frac{1}{(1-r^{2})}\,\mathbf B^{\!\top}\mathbf B\succeq0,
\;
\mathbf B:=\alpha\,\mathbf 1_{s}\mathbf 1_{q}^{\!\top}.
\]
Because $\mathbf B$ is rank 1, $\mathbf B^{\!\top}\mathbf B=\alpha^{2}s\,\mathbf 1_{q}\mathbf1_{q}^{\!\top}$ has eigenvalues $0$ and $\alpha^{2}s q$. The Schur complement~\citep{fuzhen2005schur} is therefore PSD iff its smallest eigenvalue is non-negative:
\[
(1-r^{2})-\frac{\alpha^{2}s q}{(1-r^{2})}\ge0
\Longleftrightarrow
\alpha^{2}s q\le(1-r^{2})^{2}.
\]
Hence, the mutual-information formula will take the compact form $I=-\frac12\ln(1-\varepsilon)$. In these terms, the PSD condition becomes,
\begin{equation}
0\le\varepsilon\le1,
\label{eq:PSD-bound}
\end{equation}
which is both necessary and sufficient for $\Sigma(\alpha)$ to remain PSD. At the extreme point $r=1$ ($s=q=2$) we have $d=0$, forcing $\alpha=0$ and hence $\varepsilon=0$.

\subsubsection{Trace–Squared Energy Budget}
Our maximum entropy premise (Appendix~\ref{sec:MI-derivation}) favours one global constraint on the covariance rather than ad-hoc element-wise bounds. The natural choice is the trace of the square, which measures the total mean-squared energy stored in all second moments:
\begin{equation}
\operatorname{tr}\bigl[\Sigma^{2}\bigr] =
  \sigma^{4}\Bigl[(1-r^{2})^{2}(s+q)+2\alpha^2sq\Bigr].
\label{eq:trace2}
\end{equation}
The first term accounts for $(s+q)$ node self-variances; the second captures the rank-1 inter-layer coupling energy $\varepsilon$. Then, there are exactly $(s+q)+2$ independent second-moment degrees of freedom~\citep{jaynes1957information,tipping1999probabilistic,jaynes2003probability}: $s+q$ diagonal variances and two oriented cross-layer loadings scaled by $(1-r^2)$ as in Eq.~\eqref{eq:Sigma}. Consequently, we set the budget as,
\begin{equation}
T=\sigma^{4}(1-r^{2})^{2}\bigl[(s+q)+2(1-r^2)\bigr].
\label{eq:T-budget}
\end{equation}
Choosing $T$ at its maximum-entropy value, we ensure the model cannot create additional variance (this particular budget yields the tightest PSD-compatible coupling under maximum-entropy assumptions~\citep{jaynes1957information,jaynes2003probability}); it can only redistribute the existing total between self-variances and the rank-1 cross-moments. Solving Eq.~\eqref{eq:trace2} with Eq.~\eqref{eq:T-budget} yields
\begin{equation}
\varepsilon= 1-r^2.
\label{eq:eps-from-T}
\end{equation}

\subsubsection{Tight Upper Bound on the Coupling Energy}
Because the trace budget Eq.~\eqref{eq:T-budget} fixes $\varepsilon =1-r^{2}$ while the PSD condition allows up to $\varepsilon \le 1$, then the maximal feasible value is attained at equality:
\begin{equation}
\operatorname{tr}[\Sigma^{2}]=\sigma^{4}(1-r^{2})^{2}\bigl[(s+q)+2\varepsilon\bigr], \; \varepsilon = 1-r^2.
\label{eq:eps-star}
\end{equation}
Then, the maximized $\alpha^2$ is given by,
\begin{equation}
    \alpha^2 = \frac{(1-r^2)^3}{sq}.
\end{equation}
This exhausts the trace budget; the PSD bound would allow a slightly larger $\alpha$, but it is not reachable once the energy constraint is active.
\subsubsection{Closed-Form Mutual Information}
Inserting $\varepsilon$ into the mutual-information formula
\(I_{\alpha}=-\tfrac12\ln{\bigl(1-\varepsilon\bigr)}\) gives
\begin{equation}
I_{\alpha}(\mathbf y;\mathbf x)
   =-\frac12\ln(r^{2})
   =\ln{\Bigl(\tfrac1r\Bigr)}
   =\ln{\Bigl(\tfrac{sq}{s+q}\Bigr)}.
\label{eq:MI-sq-final}
\end{equation}
Restoring $s=w_{l-1}$ and $q=w_{l}$ therefore yields the structural
layer-wise mutual information
\begin{equation}
\boxed{I(\mathbf H_{l};\mathbf H_{l-1}) \approx I_{\alpha}(\mathbf y;\mathbf x) =
   \ln{\Bigl(\frac{w_{l-1}w_{l}}
                    {w_{l-1}+w_{l}}\Bigr)}.}
\label{eq:MI-structural}
\end{equation}
The expression reveals a soft bottleneck: information flow grows sub-linearly with width and vanishes when the interface is as narrow as possible (\(r=1\), e.g.\ \(w_{l-1}=w_{l}=2\)), hence the mutual information becomes zero.

% --------------------------------------------------------------------
\subsection{Information Retention Ratio}
Since Theorem~1 provides analytic expressions for both the incremental entropy
and the entropy of every layer:
\begin{align}
H(\mathbf H_l)-H(\mathbf H_{l-1})
      &=\tfrac12
         \ln{\Bigl(nw_{l-1}(\sigma_{w_l}^{2}+\mu_{w_l}^{2})
                     \sigma_{S_l}^{2}\Bigr)},\label{eq:dH}\\[2pt]
H(\mathbf H_l)
      &=\tfrac12
         \ln{\Bigl(
           2\pi\mathrm e
           \prod_{o=1}^{l}
           nw_{o-1}(\sigma_{w_o}^{2}+\mu_{w_o}^{2})
           \sigma_{S_o}^{2}\Bigr)}.\label{eq:H}
\end{align}
Dividing Eq.~\eqref{eq:dH} by Eq.~\eqref{eq:H} we obtain the
layer-wise {information-retention ratio}
\begin{equation}
     \frac{H(\mathbf H_l)-H(\mathbf H_{l-1})}{H(\mathbf H_l)}
=\frac{\displaystyle
       \ln\!\bigl(nw_{l-1}(\sigma_{w_l}^{2}+\mu_{w_l}^{2})
                  \sigma_{S_l}^{2}\bigr)}
      {\displaystyle
       \ln(2\pi\mathrm e)+
       \sum_{o=1}^{l}
       \ln{\bigl(
           nw_{o-1}(\sigma_{w_o}^{2}+\mu_{w_o}^{2})
           \sigma_{S_o}^{2}\bigr)}}.
\label{eq:irr}
\end{equation}

% --------------------------------------------------------------------
\subsection{Effective Channel Capacity}

Multiplying Eq.~\eqref{eq:irr} by the layer-wise mutual information Eq.~\eqref{eq:MI-structural} and summing $l=1$ to $L$ gives the effective channel capacity,
\begin{equation}
\boxed{~
\phi_0
=\sum_{l=1}^{L}
\frac{\displaystyle
      \ln{\bigl(nw_{l-1}(\sigma_{w_l}^{2}+\mu_{w_l}^{2})
                 \sigma_{S_l}^{2}\bigr)}}
     {\displaystyle
      \ln(2\pi\mathrm e)+
      \sum_{o=1}^{l}
      \ln{\bigl(
          nw_{o-1}(\sigma_{w_o}^{2}+\mu_{w_o}^{2})
          \sigma_{S_o}^{2}\bigr)}}
     \ln{\bigl(\tfrac{w_{l-1}w_{l}}{w_{l-1}+w_{l}}\bigr)}}\notag
\label{eq:phi0}
\end{equation}

Furthermore, applying the same weight assumption as in Theorem~1, we arrive at the simplified implementable form of effective channel capacity,
$$\boxed{~
\phi_0
=\sum_{l=1}^{L}
\frac{\displaystyle
      \ln{\bigl(nw_{l-1}
                 \sigma_{S_l}^{2}\bigr)}}
     {\displaystyle
      \ln(2\pi\mathrm e)+
      \sum_{o=1}^{l}
      \ln{\bigl(
          nw_{o-1}
          \sigma_{S_o}^{2}\bigr)}}
     \ln{\bigl(\tfrac{w_{l-1}w_{l}}{w_{l-1}+w_{l}}\bigr)}}$$
     
In practice, if the propagation method is constant across layers and widths are non-increasing, we may approximate the above form as,
$$\boxed{~
\phi_0 \approx \sum_{l=1}^{L}
\frac{\displaystyle
      \ln{\bigl(nw_{l-1}
                 \sigma_{S_l}^{2}\bigr)}}
     {\displaystyle
      \ln(2\pi\mathrm e)+
      l\cdot
      \ln{\bigl(
          nw_{l-1}
          \sigma_{S_l}^{2}\bigr)}}
     \ln{\bigl(\tfrac{w_{l-1}w_{l}}{w_{l-1}+w_{l}}\bigr)}}\notag$$

This completes the justification of the effective channel capacity $\phi_0$.

% Next appendix: reset counter again
\setcounter{theorem}{0}
\section{Hyperparameter Configuration}\label{config}
Experiments are conducted with 2x Nvidia T4 12 GB, torch 2.2.1, torch\_geometric 2.5.0, dgl 2.2.0, SLSQP.

\begin{itemize}
    \item The dropout probability of models is searched over $[0.0, 1.0]$.
    \item We select $\text{PReLU}(\cdot)$ as the activation function~\citep{you2020design} for C$^3$E-estimated model and baselines.
    \item We select cross-entropy as the objective function for performing the semi-supervised node classification task.
    \item We select AdamW as the optimizer (SGD-based methods are optional and they can achieve even better performance, but need more time to tune). 
    \item The weight decay factors are searched from $[10^{-4}, 10^{-2}]$.
    \item The learning rates are searched from $[10^{-5}, 10^{-3}]$.
    \item The depth and hidden dimensions of baselines (non-C$^3$E models) strictly follow the original works, where the hidden dimensions and depth are searched over $w_l \in \{16, 64, 128, 256, 512, 1024, 2048\}$ and $L \in [1, 20]$.
    \item The $\eta$ is selected based on empirical results on validation sets, searched over $[0.10, 0.99]$. 
    \item The choices of $\eta$ are considered based on (i) whether optimal solutions can be generated for all baselines. (ii) $\eta$ yields shorter solution time, see Appendix~\ref{etarel} for details.
\end{itemize}

% Next appendix: reset counter again
\setcounter{theorem}{0}
\section{Sensitivity Analysis of $\eta$}
\label{etarel}

\begin{figure*}[ht]
    \centering
    \includegraphics[width=0.6\linewidth]{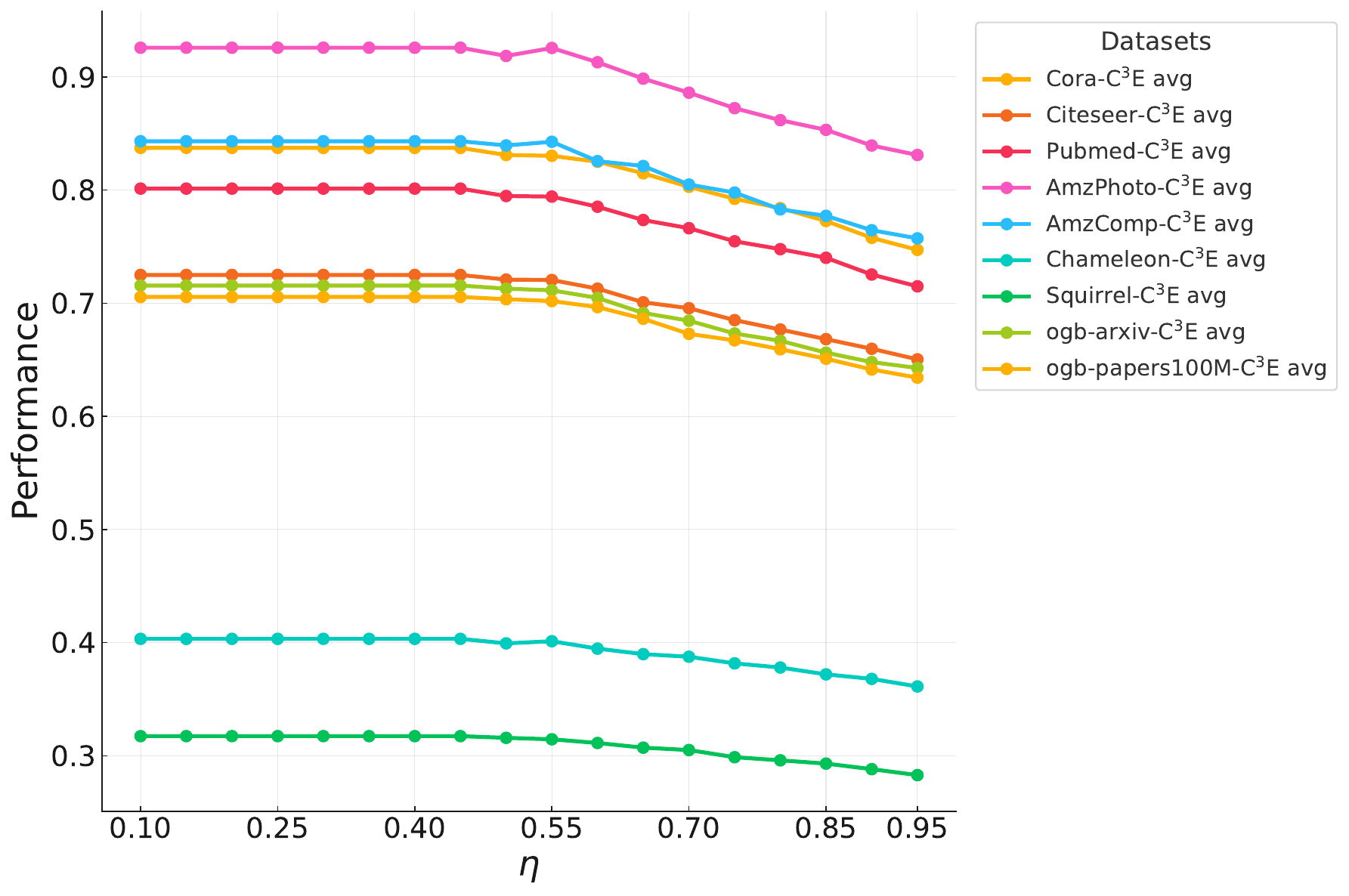}
\caption{Average performance of C$^3$E-estimated models versus $\eta$ across nine benchmark datasets. Each curve shows the averaged C$^3$E-estimated model performance per dataset.}
    \label{eta_sen}
\end{figure*}

The hyperparameter $\eta$ is a crucial regularizer within the C$^3$E framework, designed to balance the trade-off between maximizing effective channel capacity and preventing excessive information compression. To validate the robustness of our framework, we conducted a comprehensive sensitivity analysis evaluating the impact of $\eta$ on two key aspects: C$^3$E-estimated model performance and the computational cost of generating optimal solutions. Our findings, presented across Figure~\ref{eta_sen} and Table~\ref{eta-time}, reveal a clear and practical trade-off.
\begin{itemize}
    \item \textbf{Impact on Model Performance:} The analysis in Figure~\ref{eta_sen} shows that model performance is consistently high and stable across all nine datasets for $\eta$ values in the range of 0.10 to 0.45. Beyond this range, performance begins to gradually decline. This establishes a wide, robust safe region where the choice of $\eta$ has minimal impact on achieving optimal results, demonstrating that the framework does not require delicate tuning for effectiveness.
    \item \textbf{Impact on Solution Time:} The analysis in Table~\ref{eta-time} reveals an inverse relationship between $\eta$ and the time required to generate a solution. As $\eta$ increases, the computational time shrinks dramatically. However, this efficiency comes at a cost: for $\eta$ values typically greater than 0.45, the C$^3$E optimization is no longer guaranteed to find an optimal architectural solution.
    \item \textbf{Conclusion:} Taken together, these two analyses empirically validate that the C$^3$E framework is robust to the choice of $\eta$. The region $\eta \leq 0.45$ represents a reliable operating range where strong performance and guaranteed optimal solutions can be achieved, albeit at a higher computational cost. This robustness is a significant practical advantage of our method, confirming that delicate, dataset-specific fine-tuning is not a prerequisite. 
\end{itemize}

% Robustify bold command for use within S columns
\robustify\bfseries

% Set up siunitx to detect font weight
% detect-inline-weight=math ensures proper alignment in math mode (where numbers are typeset)
\sisetup{
    detect-weight=true,
    table-format=3.3, % 3 places before and 3 places after decimal point
}

%We leave the derivation of an analytical $\eta$, potentially through tighter mutual information bounds or structural graph priors, as future work.
\begin{table}[ht]
  \centering
  \resizebox{0.99 \textwidth}{!}{%
    \begin{tabularx}{\textwidth}{X *{7}{S[table-format=3.1]}}
      \toprule
      $\eta$   & {Cora} & {Citeseer} & {Pubmed} & {AmzPhoto} & {AmzComp} & {Chameleon} & {Squirrel} \\
      %\midrule
      \cmidrule(r){1-1} \cmidrule(l){2-8}
0.10  & \bfseries 88.9  & \bfseries 117.7 & \bfseries 141.8 & \bfseries 127.3 & \bfseries 137.0 & \bfseries 86.5 & \bfseries 120.1 \\
0.15  & \bfseries 78.4  & \bfseries 103.8 & \bfseries 125.0 & \bfseries 112.3 & \bfseries 120.8 & \bfseries 76.3 & \bfseries 105.9 \\
0.20  & \bfseries 60.2  &  \bfseries 79.7 & \bfseries 96.0 & \bfseries 86.2 &  \bfseries 92.7 &  \bfseries 58.6 & \bfseries 81.4 \\
0.25  & \bfseries 41.7  &  \bfseries 55.2 & \bfseries 66.5 & \bfseries 59.7 &  \bfseries 64.2 & \bfseries 40.6 & \bfseries 56.4 \\
0.30  & \bfseries 28.3  &  \bfseries 37.5 &  \bfseries 45.1 &  \bfseries 40.5 &  \bfseries 43.6 &  \bfseries 27.5 &  \bfseries 38.2 \\
0.35  & \bfseries 15.6  &  \bfseries 20.7 &  \bfseries 24.9 & \bfseries 22.3 &  \bfseries 24.0 & \bfseries 15.2 &  \bfseries 21.1 \\
0.40  &  \bfseries 7.2  &  \bfseries 9.5 &  \bfseries 11.5 &  \bfseries 10.3 &  \bfseries 11.1 &  \bfseries 7.0 &  \bfseries 9.7 \\
0.45  &  \bfseries 5.5  &   \bfseries 7.3 &   \bfseries 8.8 &   \bfseries 7.9 &   \bfseries 8.5 &   \bfseries 3.6$^*$ &  \bfseries 7.4 \\
0.50  &  \bfseries 3.7$^*$  &   \bfseries 5.4 &   \bfseries 6.2 &   \bfseries 5.3$^*$ &   \bfseries 5.7$^*$ &   3.2$^+$ &   \bfseries 5.0$^*$ \\
0.55  &  3.2$^+$  &   \bfseries 4.9$^*$ &   \bfseries 5.9$^*$ &   4.6$^+$ &   4.9$^+$ &   3.1$^+$ &   4.3$^+$ \\
0.60  &  2.9$^+$  &   3.8$^+$ &   4.6$^+$ &   4.2$^+$ &   4.5$^+$ &   2.8$^+$ &   3.9$^+$ \\
0.65  &  2.5$^+$  &   3.3$^+$ &   4.0$^+$ &   3.6$^+$ &   3.9$^+$ &   2.4$^+$ &   3.4$^+$ \\
0.70  &  2.1$^+$  &   2.8$^+$ &   3.3$^+$ &   3.0$^+$ &   3.2$^+$ &   2.0$^+$ &   2.8$^+$ \\
0.75  &  1.6$^+$  &   2.1$^+$ &   2.6$^+$ &   2.3$^+$ &   2.5$^+$ &   1.6$^+$ &   2.2$^+$ \\
0.80  &  1.3$^+$  &   1.7$^+$ &   2.1$^+$ &   1.9$^+$ &   2.0$^+$ &   1.3$^+$ &   1.8$^+$ \\
0.85  &  0.9$^+$  &   1.2$^+$ &   1.4$^+$ &   1.3$^+$ &   1.4$^+$ &   0.9$^+$ &   1.2$^+$ \\
0.90  &  0.3$^+$  &   0.4$^+$ &   0.5$^+$ &   0.4$^+$ &   0.5$^+$ &   0.3$^+$ &   0.4$^+$ \\
0.95  &  0.0$^+$ &   0.0$^+$ &   0.0$^+$ &   0.0$^+$ &   0.0$^+$ &   0.0$^+$ &   0.0$^+$ \\

      \bottomrule
    \end{tabularx}
    }%end resize box
    \vspace*{2mm}
    \caption{Average time (in seconds) for C$^3$E to generate solutions for all baselines across various datasets. The \textbf{bold} $\eta$ denotes that the optimal solutions for all baselines are generated by C$^3$E, and the symbol $^+$ denotes that for this $\eta$, C$^3$E is no longer guaranteed to generate the optimal solutions for all baselines. The symbol $^*$ denotes $\eta$ is used in the main experiments.}
      \label{eta-time}
\end{table}

% Since $\eta \in (0,1]$ is introduced to cap the effective channel capacity $\phi_0 \in [\ln{(n)}, \frac{1}{\eta}\ln{(n)}]$ and prevent uncontrolled growth of the representation compression ratio $\theta$. Determining $\eta$ involves balancing two competing factors: (i) ensuring sufficient channel capacity for information propagation with arbitrarily small loss, and (ii) avoiding over-parameterization and unstable optimization due to excessive width or depth.

% From Table~\ref{eta-time}, we can observe that as $\eta$ grows, the solution time of C$^3$E shrinks dramatically. However, as $\eta$ keeps increasing, typically beyond $\{0.45, 0.50, 0.55\}$, the optimal solutions are not guaranteed. This shows that {C$^3$E} solutions remain robust for a wide range of $\eta \leq 0.45$), indicating that $\eta$ does not require delicate fine-tuning in practice if one does not care about the solution time.
\newpage

% Next appendix: reset counter again
\setcounter{theorem}{0}
\section{Example Solutions}
\label{behavior}
\begin{figure}[hp]
  \centering
  \begin{subfigure}{\linewidth}
   \centering
    \includegraphics[width=0.7\linewidth]{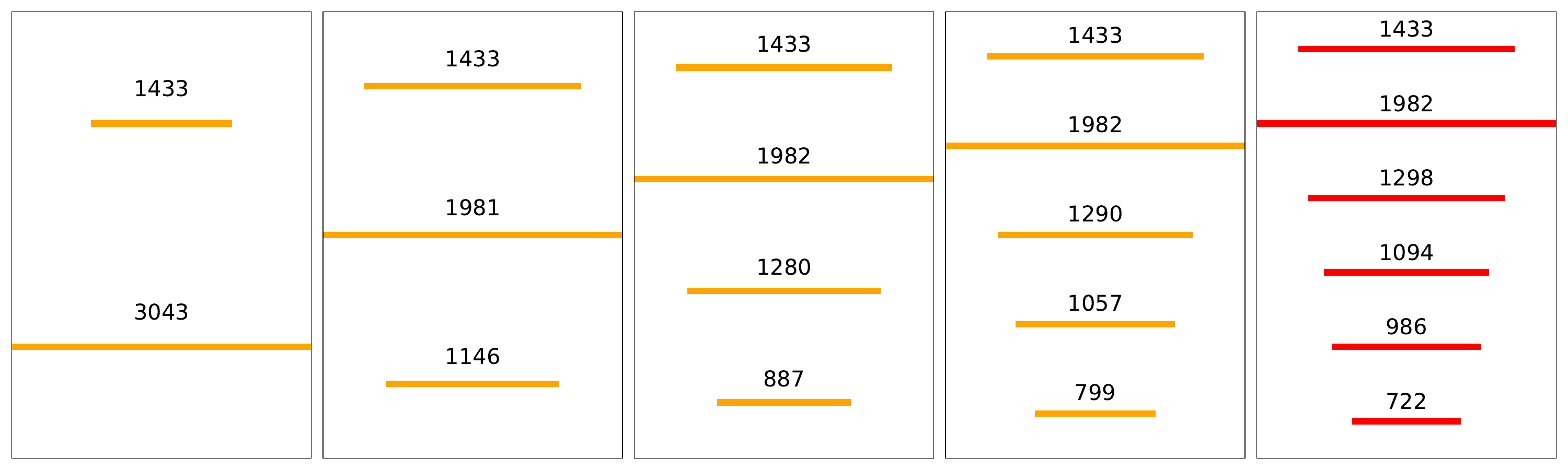}
    \caption{Cora, $n=2708$, $m=1433$.}
    \label{fig:sub1}
  \end{subfigure}\vfill
  \begin{subfigure}{\linewidth}
    \centering
    \includegraphics[width=0.7\linewidth]{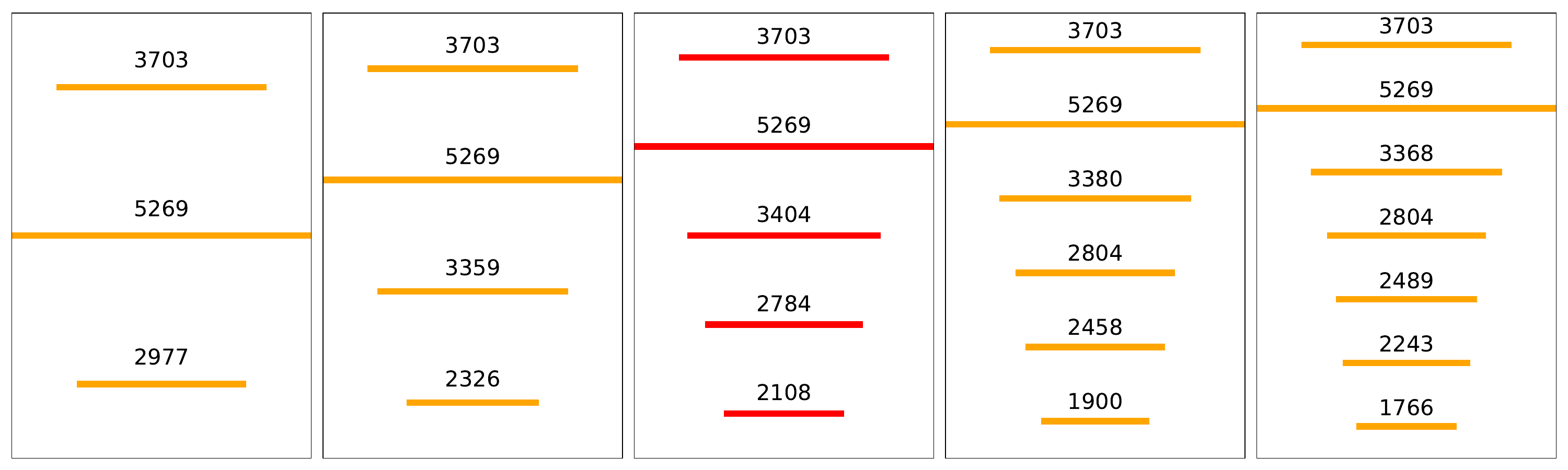}
    \caption{Citeseer, $n=3312$, $m=3703$.}
    \label{fig:sub2}
  \end{subfigure}\vfill
  \begin{subfigure}{\linewidth}
    \centering
    \includegraphics[width=0.7\linewidth]{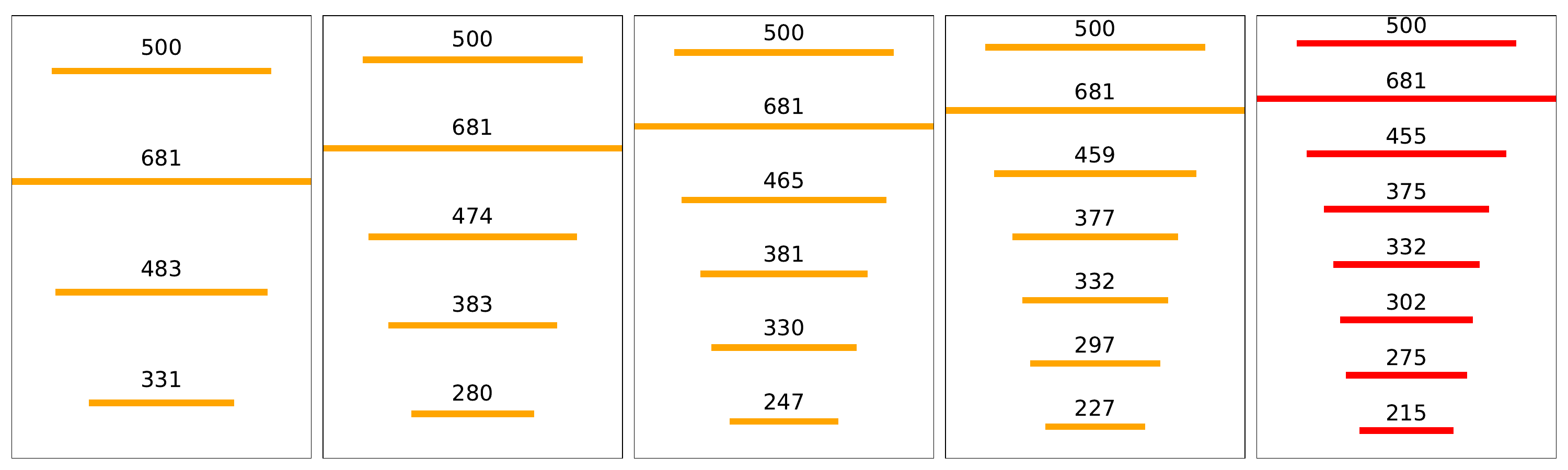}
    \caption{Pubmed, $n=19717$, $m=500$.}
    \label{fig:sub3}
  \end{subfigure}\vfill
  \begin{subfigure}{\linewidth}
    \centering
    \includegraphics[width=0.7\linewidth]{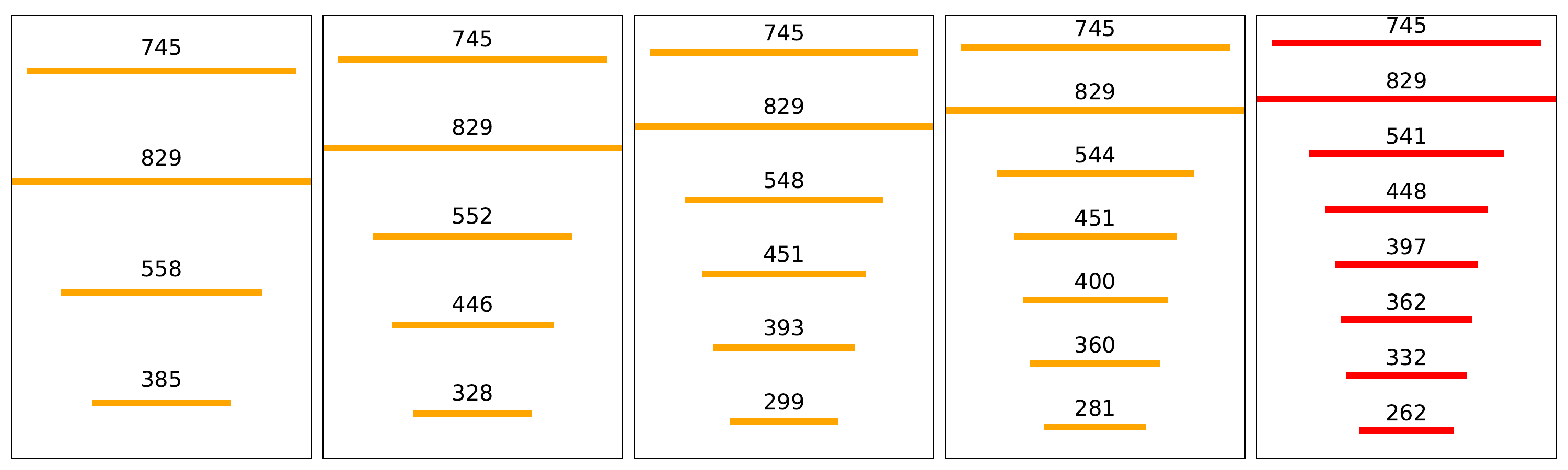}
    \caption{AmazonPhoto, $n=7650$, $m=745$.}
    \label{fig:sub4}
  \end{subfigure}\vfill
  \begin{subfigure}{\linewidth}
    \centering
    \includegraphics[width=0.7\linewidth]{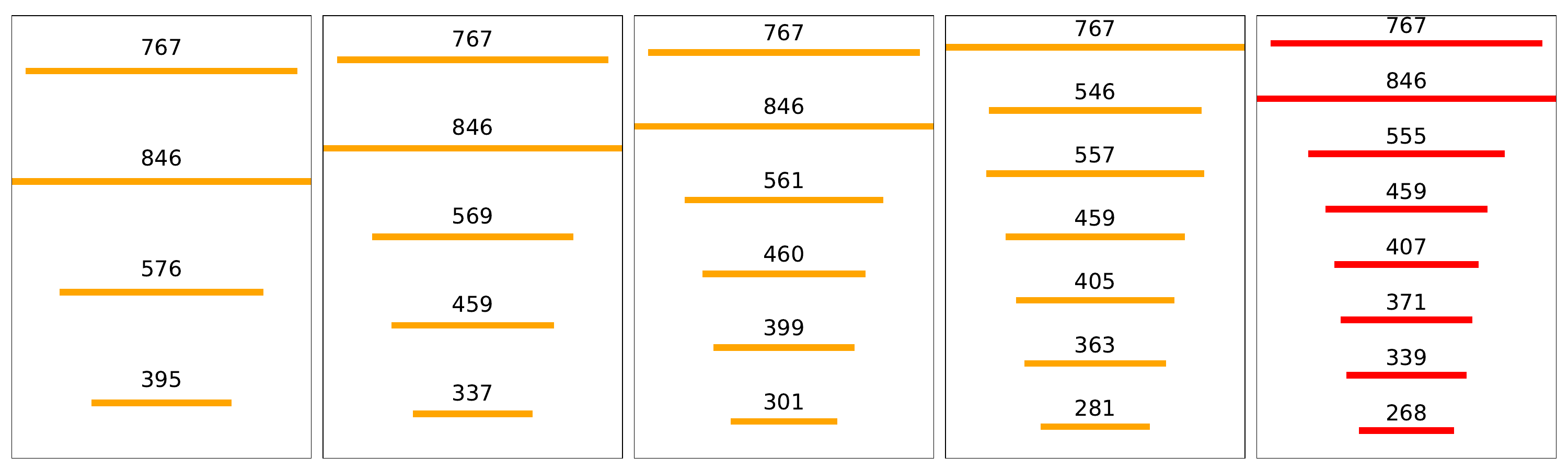}
    \caption{AmazonComputers, $n=13752$, $m=767$.}
    \label{fig:sub5}
  \end{subfigure}

  \caption{Examples of C$^3$E solutions (hidden dimensions) on five benchmark datasets.}
  \label{fig:gcn_datasets}
\end{figure}

In this section, we provide demonstrations of C$^3$E estimated hidden dimensions (optimal solutions are highlighted in red) for GCN on Cora, Citeseer, Pubmed, AmazonPhoto, and AmazonComputers. From Figure~\ref{fig:sub1} to Figure~\ref{fig:sub5}, we can observe that the solutions generated by C$^3$E consistently avoid big jumps between adjacent layers, further validating the effectiveness of C$^3$E.
\newpage

% Next appendix: reset counter again
\setcounter{theorem}{0}
\section{Visualizations of $\theta$ versus $L$}
\label{viztheta}
\begin{figure*}[tbhp]
    \centering
    \includegraphics[width=\linewidth]{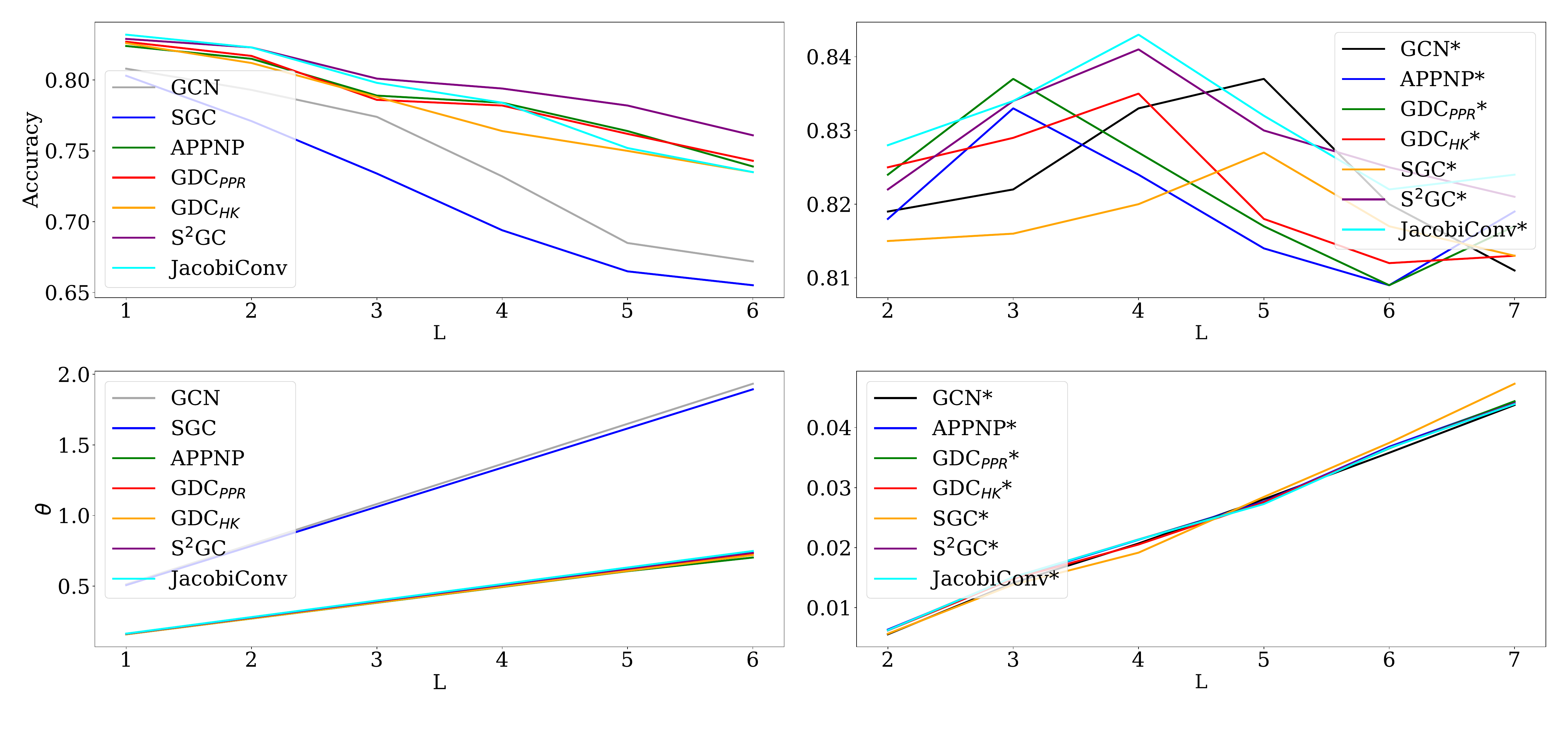}
\caption{The relationship between propagation depth ($L$), representation compression ($\theta$), and accuracy on the Cora dataset. \textbf{(Left panels)} For standard baseline models, increasing depth leads to a monotonic rise in the compression ratio $\theta$ and a consistent drop in accuracy. \textbf{(Right panels)} For {C$^3$E}-estimated models, accuracy improves with depth to an optimal point before declining, even as $\theta$ continues to rise. This highlights {C$^3$E}'s ability to find and operate within a beneficial compression regime.}
    \label{aspect_plot}
\end{figure*}

This section visualizes the relationship between propagation depth $L$, the representation compression ratio ($\theta$), and model performance on the Cora dataset, contrasting standard baselines with their C$^3$E-estimated counterparts.
\subsection{Baselines}
The results for standard baseline models are shown in the left panels.
\begin{itemize}
    \item Performance Degradation: Model accuracy consistently and monotonically decreases as the propagation depth $L$ increases from 1 to 6.
    \item Rising Compression: Concurrently, the compression ratio $\theta$ shows a monotonic and near-linear increase with depth.
\end{itemize}
These results empirically illustrate a direct correlation between rising information compression ($\theta$) and the performance degradation associated with over-squashing in standard, deep GNNs.

\subsection{{C$^3$E}-estimated models}
The right panels show a starkly different trend for models with architectures determined by {C$^3$E}.
\begin{itemize}
    \item Improved Performance with Depth: Unlike the baselines, the accuracy of {C$^3$E}-estimated models initially improves with increasing depth, reaching an optimal performance at a depth of $L\approx4$ or $L\approx5$ before declining.
    \item Beneficial Compression: This performance increase occurs even as the compression ratio $\theta$ continues to rise monotonically.
\end{itemize}
This highlights the primary finding: {C$^3$E} identifies architectures that operate within a beneficial compression regime. While deeper layers still increase compression, the {C$^3$E}-estimated models are structured to handle this compression effectively, leveraging depth to improve representation learning instead of suffering immediate degradation. This demonstrates the effectiveness of {C$^3$E} to mitigate over-squashing.

% Next appendix: reset counter again
\setcounter{theorem}{0}
\section{Post-training Representation Matrix and Learnable Weight Matrix Behavior}
\label{post-distri}
\begin{figure}[tbhp]
    \centering
    \includegraphics[width=0.45\linewidth]{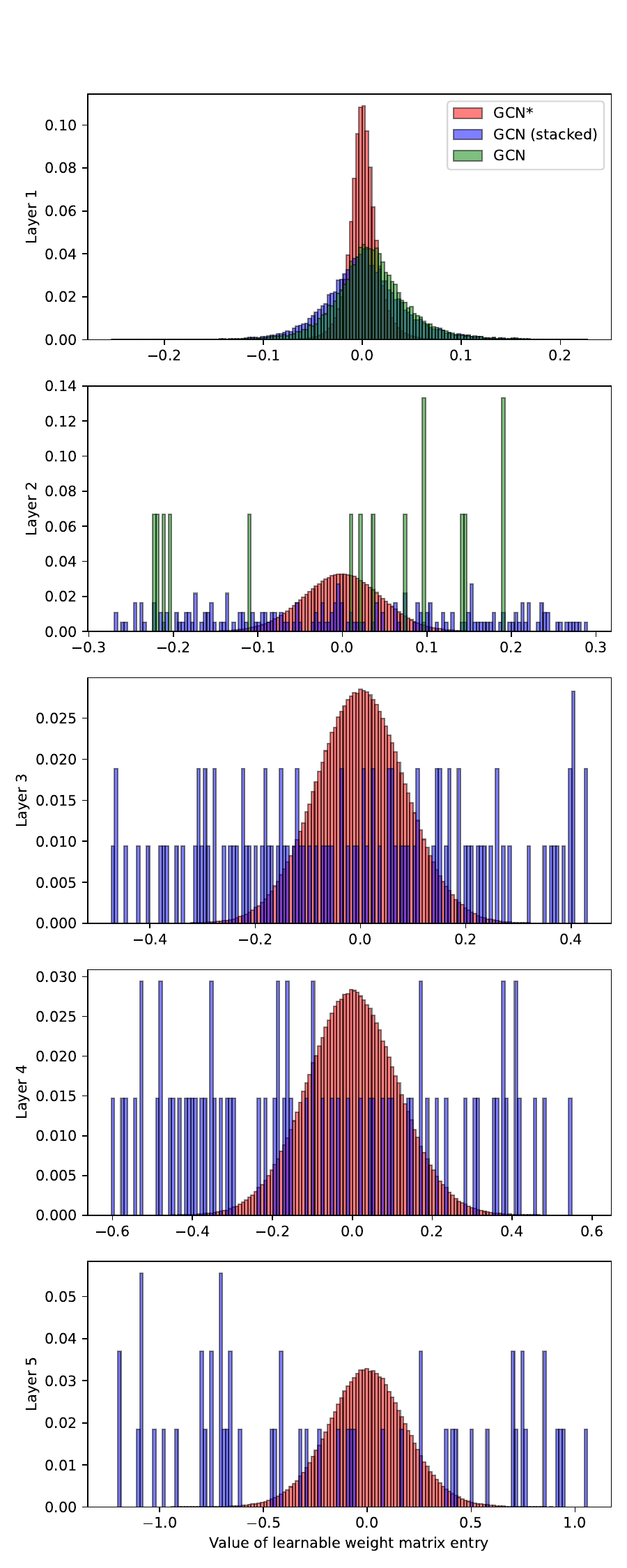}
    \caption{Learnable weight matrix entry distribution on Cora, $n=2708$, $m=1433$.}
    \label{his1w}
\end{figure}

\begin{figure}[tbhp]
    \centering
    \includegraphics[width=0.45\linewidth]{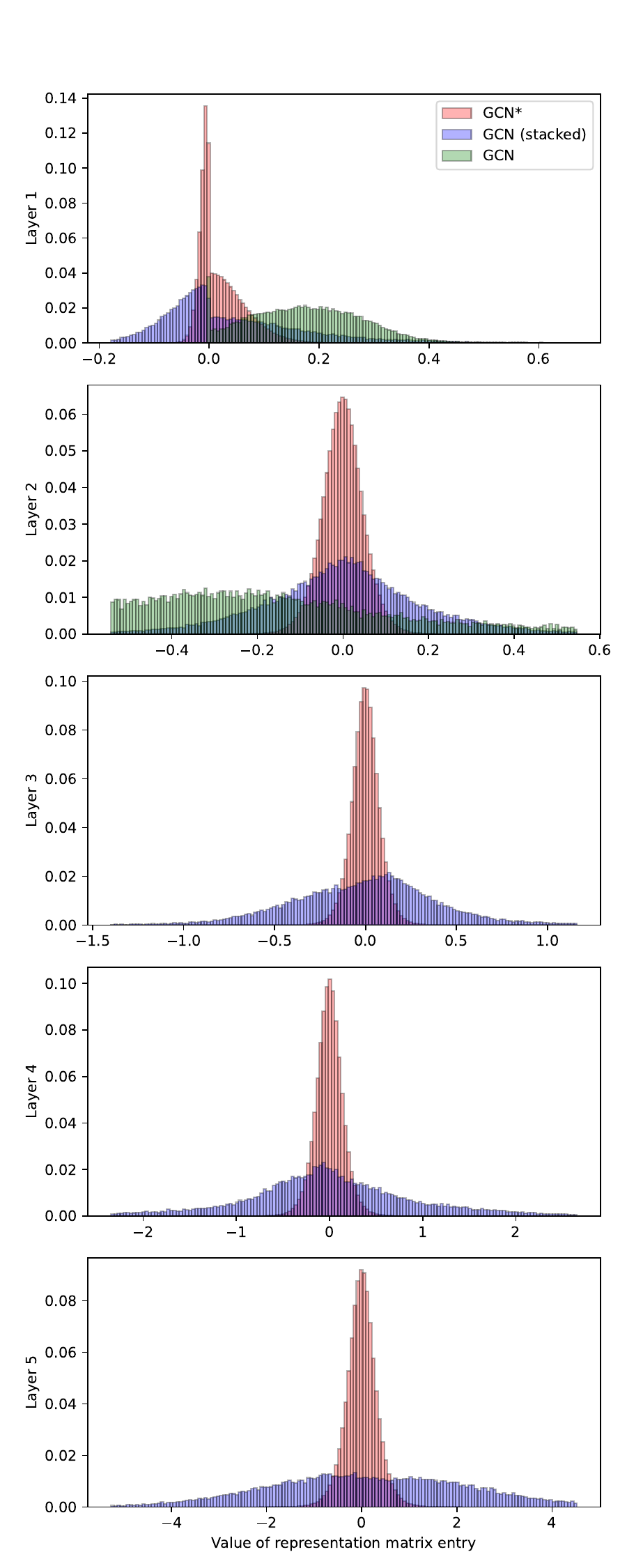}
    \caption{Representation matrix entry distribution on Cora, $n=2708$, $m=1433$.}
    \label{his1}
\end{figure}

\begin{figure}[tbhp]
    \centering
    \includegraphics[width=0.45\linewidth]{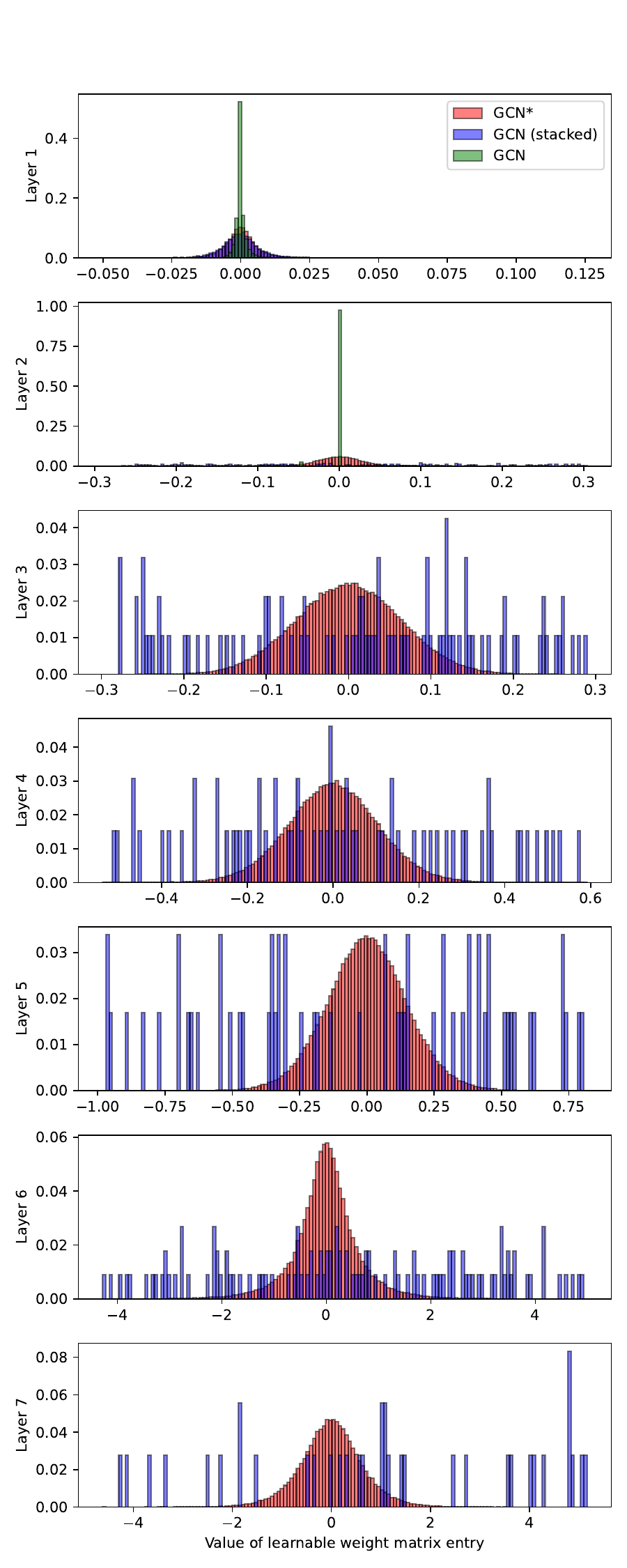}
    \caption{Learnable weight matrix entry distribution on AmazonPhoto, $n=7650$, $m=745$.}
    \label{his2w}
\end{figure}

\begin{figure}[tbhp]
    \centering
    \includegraphics[width=0.45\linewidth]{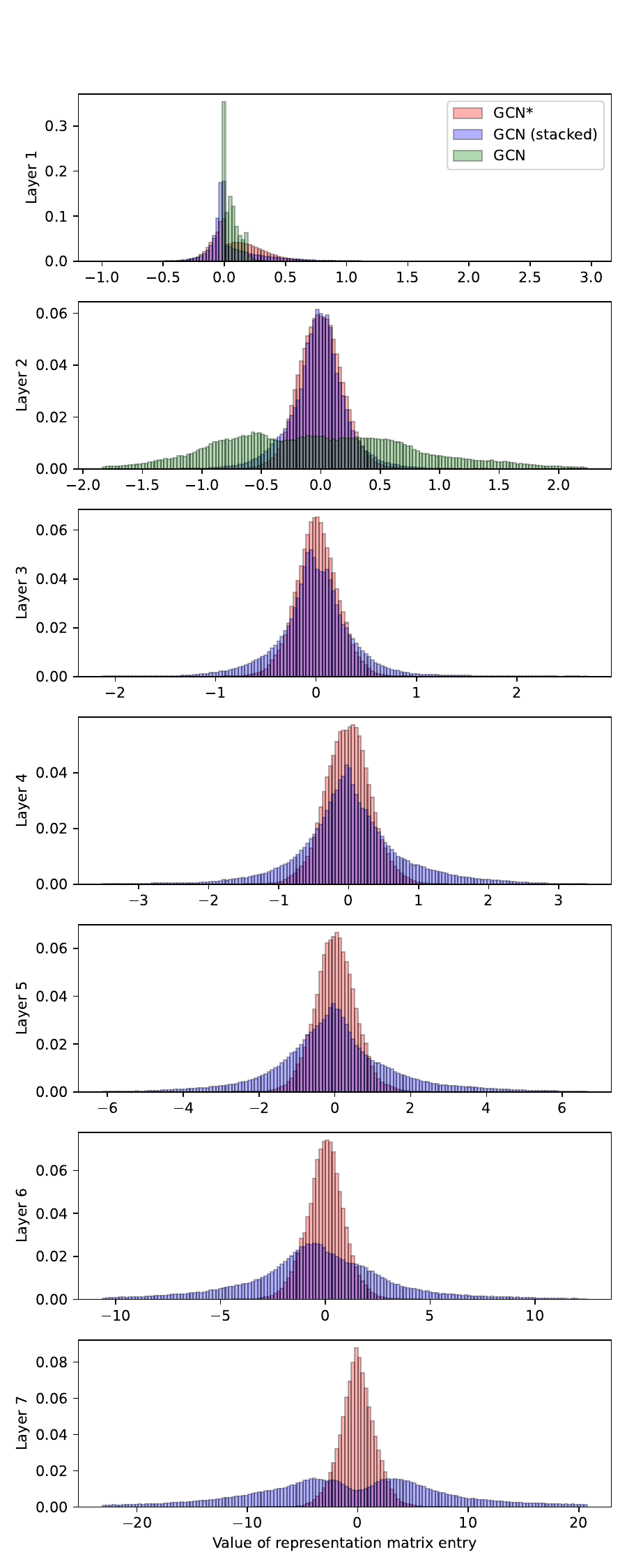}
    \caption{Representation matrix entry distribution on AmazonPhoto, $n=7650$, $m=745$.}
    \label{his2}
\end{figure}

\begin{figure}[tbhp]
    \centering
    \includegraphics[width=0.45\linewidth]{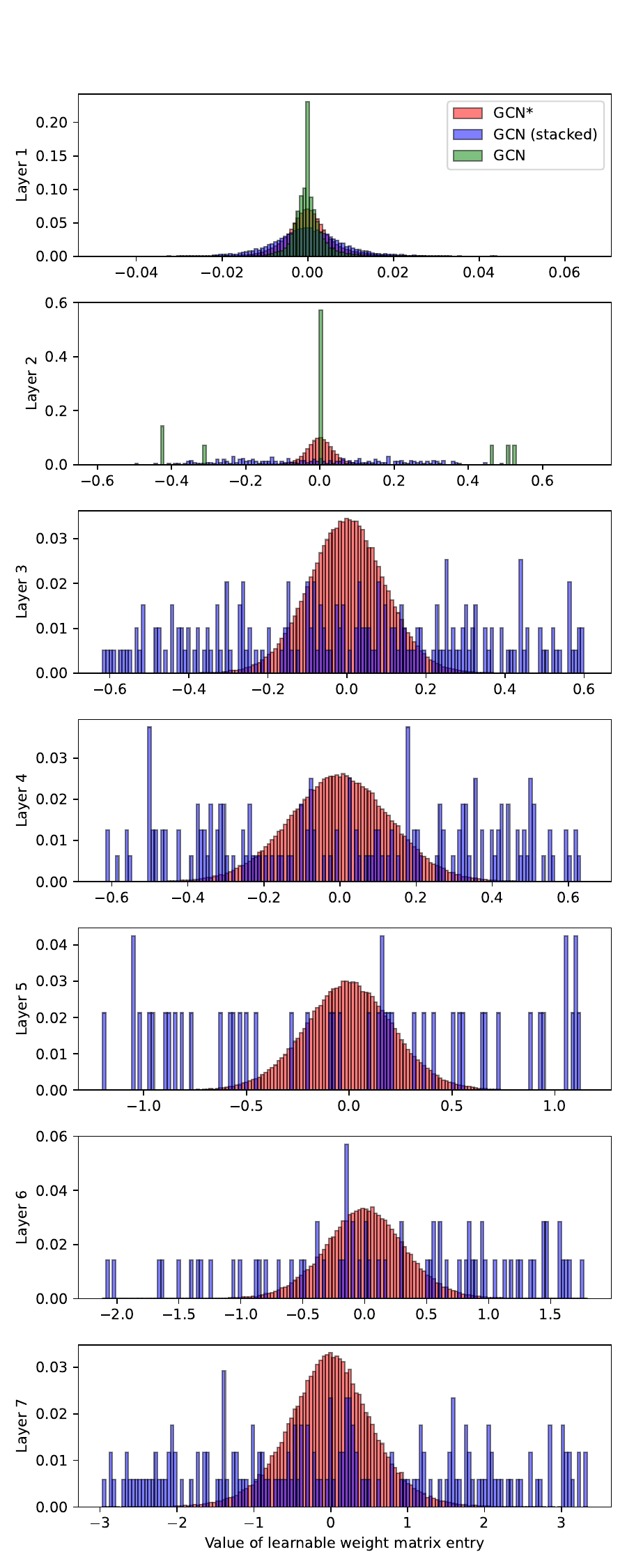}
    \caption{Learnable weight matrix entry distribution on AmazonComputers, $n=13752$, $m=767$.}
    \label{his3w}
\end{figure}

\begin{figure}[tbhp]
    \centering
    \includegraphics[width=0.45\linewidth]{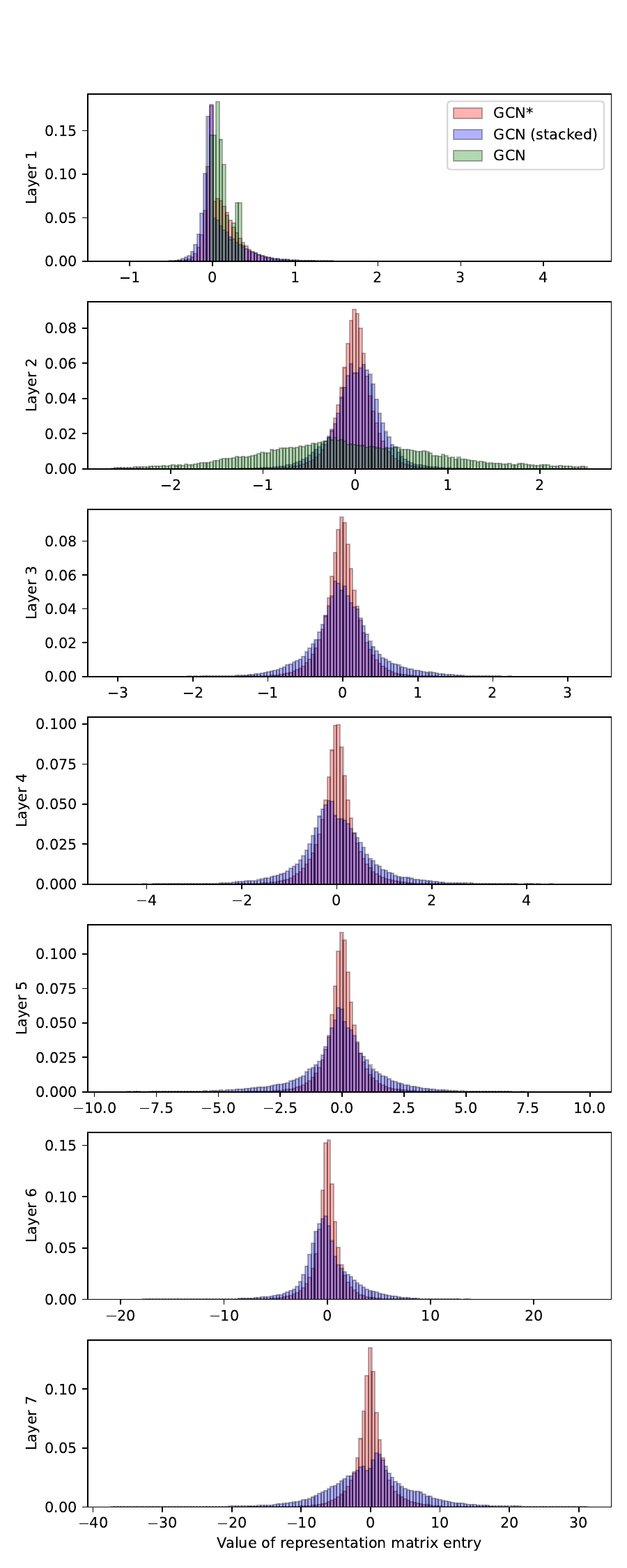}
    \caption{Representation matrix entry distribution on AmazonComputers, $n=13752$, $m=767$.}
    \label{his3}
\end{figure}

This section provides the entry-wise distribution of representation matrices and learnable weight matrices after training, visualized via histograms. In the illustrations, the y-axis denotes the probability of the entry, and the x-axis denotes the value of the entry. Based on Figure~\ref{his1w}, Figure~\ref{his2w}, Figure~\ref{his3w} for learnable weight matrix distributions, Figure~\ref{his1}, Figure~\ref{his2}, and Figure~\ref{his3} for representation matrix distributions:

\subsection{Learnable weight matrix}
\subsubsection{non-C$^3$E models}
\begin{itemize}
    \item \textbf{Channel Instability and Degeneration:} In non-C$^3$E models, the information channel quickly degenerates after the first layer. The weight distributions become highly dispersed and spiky, indicating that the network is learning an unstable and inefficient set of parameters for propagating information.
    \item \textbf{Weight range expansion during propagation:} The range of weight values (the spread along the x-axis) consistently increases after the first propagation layer, particularly for the more extreme values in the spiky distributions.
\end{itemize}
\subsubsection{C$^3$E models}
\begin{itemize}
\item \textbf{Consistent near-Gaussian shape:} The weight matrix distributions largely retain a shape resembling a Gaussian distribution across all layers. The distributions remain closely centered around zero and become progressively wider (increased variance) with depth. These suggest that each weight matrix learns meaningfully, and the overall structure avoids issues like spiky and sparse distributions or weight explosion that can be seen in models using heuristically-preferred hidden dimensions in non-C$^3$E models.
\end{itemize}

\subsection{Representation matrix}

\subsubsection{non-C$^3$E models}
\begin{itemize}
\item \textbf{Expanding activation values after layer 1:} Similar to the behavior of weight matrices, repeated propagation operations and linear transforms appear to amplify certain eigenmodes. The post-activation representation entries cover a wider range of values as the propagation depth increases, indicating a potential for exploding activations.
\item \textbf{Information Loss and Feature Flattening:} The representation matrices exhibit uncontrolled expansion, and their distributions progressively flatten. This demonstrates a tangible loss of distinctive information; as features become less concentrated, the representation channel is failing to preserve the specific, salient signals required for the downstream task.
\end{itemize}
\subsubsection{C$^3$E models}
\begin{itemize}
\item \textbf{Stable Information Flow:} The C$^3$E models maintain a stable information flow. The representation distributions exhibit a controlled variance and remain well-behaved, avoiding both exploding and vanishing activations. This indicates that the channel capacity is appropriately matched to the information being processed, allowing for effective propagation through deep layers~\citep{glorot2010understanding}.
\end{itemize}

\subsection{Overall insights}
(i) Simply scaling up with more layers without appropriately selecting hidden dimensions (as seen in stacked baselines) can lead to learnable weight distributions that become sparse and spiky after the initial layer; (ii) Properly balancing the hidden dimensions and depth, as achieved by C$^3$E creates a stable and efficient information channel. This channel supports controlled weight and activation distributions throughout the network, a condition that is essential for mitigating information loss and enabling meaningful representation learning in deep GNNs~\citep{glorot2010understanding,roberts2022principles}.




% End of appendix

\end{document}